\documentclass[aoas,MSNbibl,nameyear,rotating,dvips]{arximspdf}
\usepackage{pseudocode,multirow,dcolumn}
\usepackage{stfloats}
\usepackage{graphicx}

\doi{10.1214/13-AOAS712} 
\volume{8}
\issue{2}
\pubyear{2014}
\firstpage{926}
\lastpage{955}

\makeatletter
  \fnbelowfloat
\newcolumntype{d}[1]{D{.}{.}{#1}}

\def\cal{\mathcal}
\newcommand{\comments}[1]{}
\makeatother

\begin{document}
\begin{frontmatter}

\title{MONEYB\lowercase{a}RL: Exploiting pitcher decision-making using~Reinforcement~Learning}
\runtitle{MONEYB\lowercase{a}RL: Exploiting baseball pitcher decision-making}

\begin{aug}
\author[A]{\fnms{Gagan} \snm{Sidhu}\corref{}\ead[label=e1]{gagans@cs.ualberta.ca}\thanksref{t1,naujas}\ead[label=e1]{gagan@g-a.ca}\ead[label=u1,url]{http://webdocs.cs.ualberta.ca/\textasciitilde gagans}}
\and
\author[B]{\fnms{Brian} \snm{Caffo}\ead[label=e2]{bcaffo@gmail.com}\thanksref{t2}\ead[label=e3]{bcaffo@jhsph.edu}\ead[label=u2,url]{http://www.biostat.jhsph.edu}}

\runauthor{G. Sidhu and B. Caffo}
\affiliation{General Analytics\thanksmark{t1}\thanksref{f1}, University of Alberta\thanksmark{naujas}
and Johns Hopkins University\thanksmark{t2}}

\thankstext{f1}{\protect\url{http://www.g-a.ca}.}
\address[A]{University of Alberta\\
1-36 Athabasca Hall\\
Edmonton, Alberta T6G2E8\\
Canada\\
\printead{e1}\\
\printead{u1}}
\address[B]{Department of Biostatistics\\
Johns Hopkins University\\
615 N. Wolfe Street\\
Baltimore, Maryland 21205\\
USA\\
\printead{e3}\\
\printead{u2}}
\end{aug}

\received{\smonth{9} \syear{2012}}
\revised{\smonth{12} \syear{2013}}

%
\begin{abstract}
This manuscript uses machine learning techniques to exploit baseball
pitchers' decision making, so-called ``Baseball IQ,'' by modeling
the at-bat information, pitch selection and counts, as a Markov
Decision Process (MDP). Each state of the MDP models the pitcher's
current pitch selection in a Markovian fashion, conditional on the
information immediately prior to making the current pitch. This
includes the count prior to the previous pitch, his ensuing pitch
selection, the batter's ensuing action and the result of the pitch.

The necessary Markovian probabilities can be estimated by the
relevant observed conditional proportions in MLB pitch-by-pitch
game data. These probabilities could be pitcher-specific, using
only the data from one pitcher, or general, using the data from a
collection of pitchers.

Optimal batting strategies against these estimated conditional
distributions of pitch selection can be ascertained by Value
Iteration. Optimal batting strategies against a pitcher-specific
conditional distribution can be contrasted to those calculated
from the general conditional distributions associated with a
collection of pitchers.

In this manuscript, a single season of MLB data is used to calculate
the conditional distributions to find optimal pitcher-specific and
general (against a collection of pitchers) batting strategies. These
strategies are subsequently evaluated by conditional distributions
calculated from a different season for the same pitchers. Thus, the
batting strategies are conceptually tested via a collection
of simulated games, a ``mock season,'' governed by distributions not
used to create the strategies. (Simulation is not needed, as exact
calculations are available.)

Instances where the pitcher-specific batting strategy outperforms the
general batting strategy suggests that the pitcher is
\textit{exploitable}---knowledge of the conditional distributions of
their pitch-making decision process in a different season yielded a
strategy that worked better in a new season than a general batting
strategy built on a population of pitchers. A permutation-based
test of exploitability of the collection of pitchers
is given and evaluated under two sets of assumptions.

To show the practical utility of the approach, we introduce a
spatial component that classifies each pitcher's pitch-types using a
batter-parameterized spatial trajectory for each pitch. We found
that heuristically labeled ``nonelite'' batters benefit from using
the exploited pitchers' pitcher-specific strategies, whereas (also
heuristically labeled) ``elite'' players do not.
\end{abstract}



\end{frontmatter}

\section{Introduction}\label{sect:intro}
\begin{quote}
\textit{``Good pitching will always stop good hitting and vice-versa.''}
\end{quote}
\begin{flushright}
\footnotesize---Casey Stengel
\end{flushright}\vspace*{6pt}

Getting a hit off of a major league pitcher is one of the hardest tasks
in all of sports.
Consider the fact that a batter in possession
of detailed knowledge of a~pitcher's processes for determining pitches to
throw would
have a large advantage for exploiting that pitcher to get on
base [\citet{bbcoach}]. Pitchers apparently reveal an enormous amount of
information regarding their behaviour through their historical
game data [\citet{tap}]. However, making effective use of this data is challenging.

This manuscript uses statistical and machine learning techniques to:
(i) represent
specific pitcher and general pitching behaviour by Markov processes
whose transition probabilities
are estimated, (ii) generate
optimal batting strategies against these processes, both in the general and
pitcher-specific sense, (iii) evaluate those strategies on data not used
in their creation, (iv) investigate the implication of the strategies on
pitcher exploitability, and (v) establish the viability of the use of
algorithmically/empirically-derived batting strategies in real-world
settings. These goals are accomplished by a detailed analysis of two
seasons of US Major League Baseball pitch-by-pitch data.

Parsimony assumptions are necessary to appropriately
represent a pitcher's behaviour. For each pitcher, it is assumed that
their pitch behaviour
is stochastic and governed by one-step Markovian assumptions on the
pitch count.
Specifically, each of the twelve unique nonterminal states of the pitch
count are modeled
as a Markov process.
It is further assumed that the relevant transition probabilities can be
estimated
by data of observed pitch selections at the twelve unique
pitch counts. It should be noted that the transition could be estimated
for a particular pitcher based on their historical data, or historical
data for a representative
collection of pitchers to investigate general pitching behaviour.

It is herein demonstrated that a pitcher's decisions can be
exploited by a data-informed batting strategy that takes
advantage of their mistakes at each pitch count in the at-bat. To
elaborate, if
indeed a pitcher's behaviour is well modeled by a Markov process
on the pitch count, a batter informed of the relevant transition probabilities
is presented with a Markov Decision Process (MDP) to swing or stay at
a given pitch count. Optimal batting strategies for MDPs
can be found by a Reinforcement Learning (RL) algorithm. RL is a subset of
artificial intelligence for finding (Value Iteration) and evaluating
(Policy Evaluation) optimal strategies in stochastic
settings governed by Markov processes. It has been used successfully in
sports gamesmanship,
via the study of offensive play calling in American football [\citet
{ndp}]; see Section \ref{sect:ndpdisc} for further discussion.
The RL Value Iteration algorithm applied to the pitcher-specific
Markov transition probabilities yields a pitcher-specific
batting strategy. In the event that the Markov transition probabilities
were estimated from a
representative collection of pitchers, a general optimal batting
strategy would result from RL Value Iteration.

\begin{figure}

\includegraphics{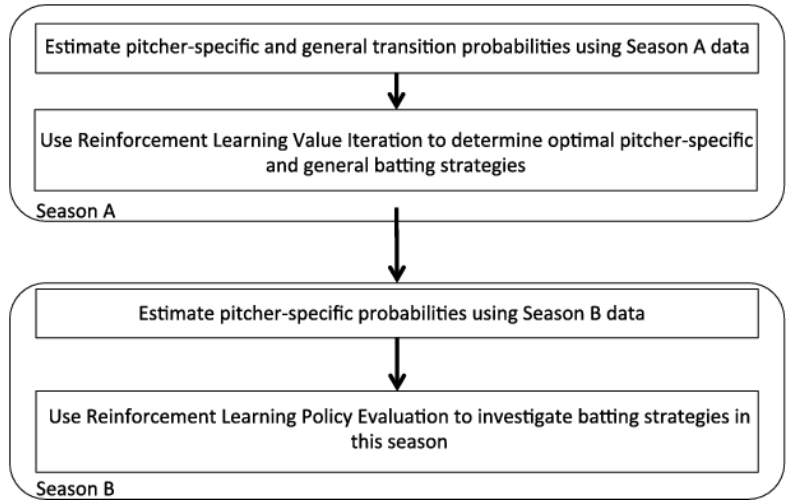}

\caption{A schematic of the Reinforcement Learning Value Iteration and
Policy Evaluation analysis
of two seasons of MLB data.}\label{fig:diagram}
\end{figure}

An important component of the development of optimal batting strategy
is their
evaluation. To this end, RL Policy Evaluation is used to investigate
the performance of
batting strategies on pitcher-specific and general Markov transition
probabilities
estimated from data not used in the RL Value Iteration algorithm to develop
the strategies. A schematic of the analysis pipeline is given in Figure \ref{fig:diagram}.

In addition to evaluating the batting strategies on new data,
comparison of the performance of the
pitcher-specific and general batting strategies yields important
information on the utility of
an optimal, data-informed batting strategy against a specific pitcher.
To this point,
a pitcher has been \textit{exploited} if Policy Evaluation suggests
that the
pitcher-specific optimal batting strategy against them is superior to
the general optimal batting strategy, with their
difference or ratio estimating the degree of exploitability. The
term ``exploited'' is used in the sense that an opposing batter would
be well served in carefully
studying that pitcher's historical data, rather than executing a
general strategy.

Given this framework, it is possible to investigate hypotheses on
general pitcher exploitability
using permutation tests.
However, the nature of the tests requires assumptions on the direction
of the alternative.
Under the assumption that all pitchers are not equally exploitable, the
hypothesis
that pitchers can be exploited more than 50\% of the time can be investigated.
This hypothesis was not rejected at a 5\% error rate. Under the
assumption that all
pitchers are equally exploitable, the hypothesis was rejected. It is
our opinion that the assumption of unequal exploitability is better
suited for baseball's at-bat setting (see Section \ref{sect:comps}).
Thus, this manuscript is the first to provide statistical evidence in
support of strategizing against specific pitchers instead of a group of
pitchers.

\begin{figure}

\includegraphics{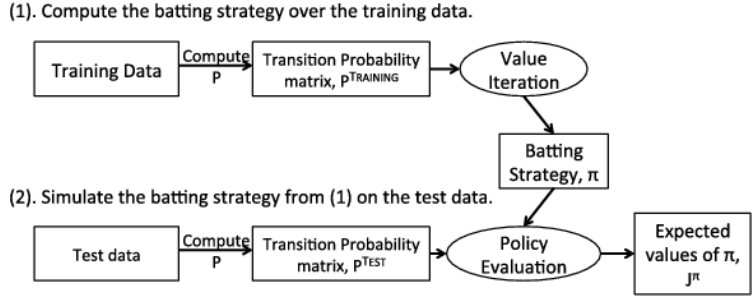}

\caption{Illustration of the two-stage process of computing and
evaluating the batting strategy that is computed over the input
training data. This strategy is then given as input, along with the
test data's transition probability matrix $P^{\textsc{\scriptsize Test}}$, to the
Policy Evaluation algorithm, which outputs the expected rewards for
each state when following batting strategy~$\pi$.}
\label{fig:strategy}
\end{figure}

To highlight the utility of an optimal pitcher-specific batting
strategy for an exploitable pitcher,
a data-driven validation was conceived. This validation associates
spatial trajectories with the pitch-type that
was employed. The spatial component is a classifier that estimates the
pitch-type of a batter-parameterized spatial trajectory after training
over the respective pitcher's actual pitch trajectories. The estimated
pitch-type selects the batting action from the exploited pitcher's
pitcher-specific batting strategy (Section \ref{sect:stratspat};
the schematic diagram outlining this process is provided in
Figure \ref{fig:strategy}). The
simulation therefore uses the spatial and strategic component to
realistically simulate a batter's performance when facing an exploited
pitcher. The batter's actual and simulated statistics when facing the
respective pitcher are then compared using typical baseball statistics.

It was found that heuristically-labeled ``elite'' batters'
simulated statistics are worse than their actual statistics. However,
it was also found that the (also heuristically-labeled)
``nonelite''\setcounter{footnote}{1}\footnote{In our study, nonelite batters are excellent
players, some having participated
in the MLB All-Star game.} batters' simulated statistics were greatly
improved from their actual statistics. The simulation results suggest
that an exploited pitcher's pitcher-specific strategies are useful for
nonelite batters.

The manuscript is laid out as follows: The section immediately
following this paragraph provides a very brief demonstration of
Reinforcement Learning algorithms to help stimulate understanding,
Section \ref{sect:ndp} discusses the strategic component, which applies
Reinforcement Learning algorithms to Markov processes to compute \textit
{and} evaluate the respective batting strategies, and Section \ref{spatdesc} discusses the spatial component, which simulates a specific
batter's performance using an exploited pitcher's pitcher-specific
batting strategy. Section \ref{methods} provides the methodology used
to produce the results given in Section \ref{results}.

\begin{figure}

\includegraphics{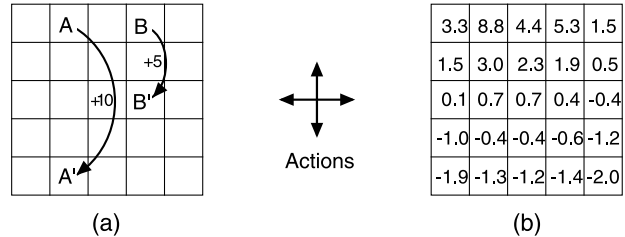}

\caption{GridWorld. The values in the right grid are determined using
Policy Evaluation with a policy that randomly selects one of the four
actions at each state. Reprinted with permission from Richard S. Sutton
and Andrew G. Barto, Reinforcement Learning: An Introduction, published
by The MIT Press.}\label{fig:gw}
\end{figure}

\subsection*{Reinforcement Learning tutorial}\label{sect:toy}
Reinforcement Learning focuses on the problem of \textit
{decision-making facing uncertainty}, which are settings where the
decision-maker (agent) interacts with a new, or unfamiliar,
environment. The agent continually interacts with the environment by
selecting actions, where the environment then responds to these actions
and presents new scenarios to the agent [\citet{sb}]. This environment
also provides \textit{rewards}, which are numerical values that act as
feedback for the action selected by the agent in the environment. At
time-unit \textit{t}, the agent is given the environment's state $s_t
\in\mathcal{S}$, and selects action $a_t \in\mathcal{A}(s_t)$, where
$\mathcal{A}(s_t)$ is the set of all possible actions that can be taken
at state $s_t$. Selecting this action increments the time-unit, giving
the agent a reward of $r_{t+1}$ and also causing it to transition to
state $s_{t+1}$. Reinforcement Learning methods focus on how the agent
changes its decision-making as a consequence of its experiment in the
respective environment. The agent's goal is to use its knowledge of the
environment to maximize its reward over the long run. Specifying the
environment therefore defines an instance of the Reinforcement Learning
problem [\citet{sb}] that can be studied further.

\textit{GridWorld} (Figure \ref{fig:gw}) is a canonical example that
illustrates how Reinforcement Learning algorithms can be applied to
various settings. In this example, there are four equiprobable actions
that can be taken at each square (state): left, right, up and down,
where each action is selected at random Any action taken at either
square A or B yields a reward of $+10$ and $+5$, respectively, and
transports the user to square A$^\prime$ or B$^\prime$, respectively.
For all other squares, a reward of 0 is given for actions that do not
result in falling off the grid, where the latter outcome results in a
reward of~$-1$. 

The negative values in the lower parts of the grid demonstrate
that the expected reward of square A is below its immediate reward
because after we are transported to square A$^\prime$, we are likely to
fall off the grid. Conversely, the expected reward of square B is
higher than its immediate reward because after we are transported to
square B$^\prime$, the possibility of running off the grid is
compensated for by the possibility of running into square A or B [\citet{sb}].

Employing Reinforcement Learning algorithms in various real-world
settings allows us to ``balance'' the immediate and future rewards
afforded by the outcomes at each state with their respective
probabilities. This approach enables the computation of a sequence of
actions, or \textit{policy}, that maximize the immediate reward while
considering the consequences of these actions.

\section{Strategic component---Reinforcement Learning in baseball
(RLIB)}\label{sect:ndp}

\subsection{Markov processes}\label{subsect:markov}

\begin{figure}

\includegraphics{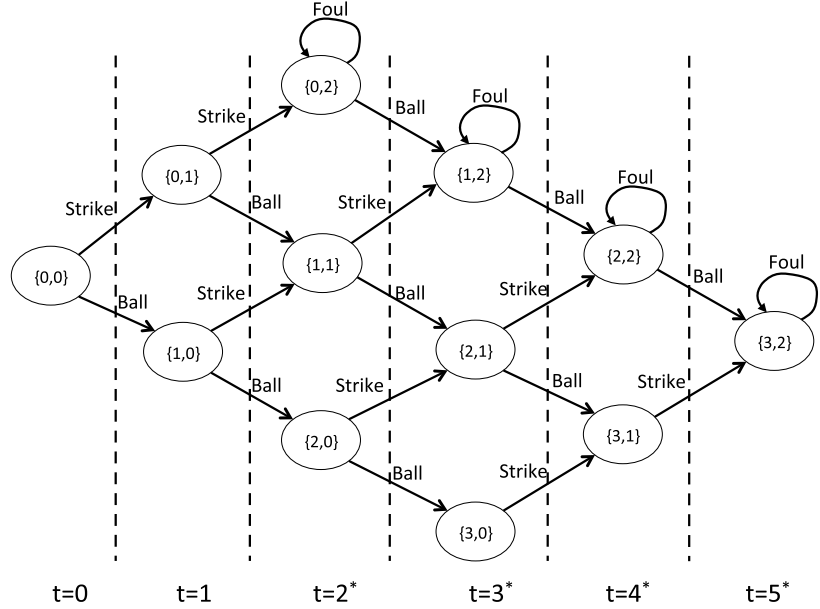}

\caption{Our view of an at-bat as a Markov process, where each oval
denotes the state $\{B,S\}$ in the at-bat, where $B$ and $S$ are the number
of balls and strikes, respectively, arrows represent transitions, and $t$
is the time unit that is the number of pitches thrown in the respective
at-bat. Asterisks denote that some states at the respective time-unit
are still valid transitions at higher time-units---that is, if a batter
fouls off a pitch at $t=3$ in the $\{1,2\}$ pitch count, they will still be
in this state at $t=4$. We omit the terminal (absorbing) states for neatness.}
\label{fig:markovrep}
\end{figure}

Let $\{X^{(t)} \}$ be a Markov process with finite state space ${\cal
S} = \{E_1, \ldots, E_n\}$ where the states represent pitch counts. We
assume stationary transition probabilities. That is, $p_{E_k
\rightarrow E_j} = P(X^{(t)} = E_j  |  X^{(t-1)} = E_{k})$ for $E_j,
E_k \in{\cal S}$ is the same for all $t$ [\citet{feller}]. Figure \ref{fig:markovrep} displays the Markov transition diagram omitting the
absorbing states (hit, out and walk).

We define optimal policies as a set of actions that maximize the
expected reward at every state in a Markov process. Conditioning on a
batter's action at a state yields the probability distribution for the
immediate and future state. Since the at-bat always starts from the
$s_0 = \{0,0\}$ state, the long-term reward is defined as
%
\begin{equation}
\label{eqn:vf} J^*(s_0) = \max_{\pi= \{u^0,\ldots, u^T \}} \mathbb{E}
\Biggl\{\sum_{t=0}^T g\bigl(s_t,
u^t,s_{t+1}\bigr) \Big| s_0 = \{0,0\}, \pi \Biggr\},
\end{equation}
where:
\begin{itemize}
\item$\{s_t\}$ is the sequence of states, or pitch counts, in the state
space $\mathcal{S}$ visited by the batter for the respective at-bat. We
define the set of \textit{terminal states}---that is, states that
conclude the at-bat---as $\mathcal{E} = \{\mathrm{O},\mathrm{S},\mathrm
{D},\mathrm{T},\break\mathrm{HR}, \mathrm{W}\} \subset\mathcal{S}$, where
\textup{O, S, D, T, HR, W} are abbreviations for Out, Single, Double,
Triple, Home Run and Walk, respectively.
\item$\pi= \{u^0,\ldots, u^T\}$ is the best batting strategy that
contains the batting actions that maximize the expected reward of every
state. Since the batter can Swing or Stand at each state, it follows
that $u= \mathrm{Swing}$ or $u= \mathrm{Stand}$ for each nonterminal
state.
%
\item$g(s_t,u^t,s_{t+1})$ is the reward function whose output reflects
the batter's preference of transitioning to state $s_{t+1}$ when
selecting the best batting action $u^t$ from state $s_t$. The reward
function used in our study is
\[
g(i,u,j) = \cases{ %
0, &\quad$\mbox{if \textit{j}${}={}$\{O\} or $j
\in\mathcal{S}\cap\mathcal {E}^c$},$\vspace*{2pt}
\cr
1, &\quad$\mbox{if
\textit{j}${}={}$\{W\} and \textit{u}${}={}$Stand},$ \vspace*{2pt}
\cr
2, &\quad$
\mbox{if \textit{j}${}={}$\{S\} and \textit{u}${}={}$Swing},$ \vspace*{2pt}
\cr
3,
&\quad$\mbox{if \textit{j}${}={}$\{D\} and \textit{u}${}={}$Swing},$
\vspace*{2pt}
\cr
4, &\quad$\mbox{if \textit{j}${}={}$\{T\} and
\textit{u}${}={}$Swing},$ \vspace*{2pt}
\cr
5, &\quad$\mbox{if \textit{j}${}={}$
\{HR\} and \textit{u}${}={}$Swing}$ } \qquad\forall i \notin\mathcal{E}.
\]
Further information on our formulation of the reward function can be
found in Section \ref{sect:stratdisc}.
\end{itemize}

Equation (\ref{eqn:vf}) can be viewed as the maximized expected
reward of state $\{0,0\}$ when following batting strategy $\pi$, which
is comprised of actions that maximize the expected reward over all of
the at-bats $\{s_t\}$ given as input [\citet{ndpref}].

To find the batting actions that comprise the best batting strategy, we
find the batting action $u$ that satisfies the optimal expected reward
function for state $i$ [\citet{ndpref}]:
%
\begin{equation}
\label{eqn:opvalstate} J^*(i) = \max_{u \in U} \biggl[ \sum
_{j \in\mathcal{S}} P_u^{\textsc
{\scriptsize{Training}}}(i,j) \bigl(g(i,u,j) +
J^*(j)\bigr) \biggr]\qquad \forall i \in \mathcal{S}\cap\mathcal{E}^c,
\end{equation}
where $P_u^{\textsc{\scriptsize Training}}(i,j)$ is an estimated probability of
transitioning from state \textit{i} to \textit{j} when selecting
batting action \textit{u} on the pitch-by-pitch data 
and $J^*(i)$ is the maximized expected
reward of state $i$ when selecting the action $u$ that achieves this maximum.

The Value Iteration algorithm, shown in Figure \ref{fig:valueiteration},
solves for the batting actions that satisfy
equation (\ref{eqn:opvalstate}). Intuitively, the algorithm keeps
iterating until the state's reward function is close to its optimal
reward function $J^*$, where in the limit $J^* = \lim_{k\to\infty
}J^k(i)\   \forall i \in\mathcal{S}$ [\citet{ndp}]. The algorithm
terminates upon satisfying the convergence criterion $\Delta< \varepsilon
$, where $\varepsilon= 2.22\times10^{-16}$ is the machine-epsilon
predefined in the MATLAB programming language; this epsilon ensures
that the reward functions of each state have approximately 15--16 digits
of precision.

The Policy Evaluation algorithm, shown in Figure \ref{fig:pealg},
uses the best batting strategy $\pi$ on a different season of the
respective pitcher's pitch-by-pitch data 
to calculate the expected reward of each state in the
at-bat, given by
%
\begin{equation}
\label{eqn:peeq} J^\pi(i) = \sum_{j \in\mathcal{S}}
P_{\pi(i)}^{\textsc{\scriptsize Test}}(i,j) \bigl[g\bigl(i,\pi(i),j\bigr) +
J^\pi(j) \bigr]\qquad \forall i \in\mathcal{S} \cap
\mathcal{E}^c,
\end{equation}
where 
$J^\pi(i)$ is the
expected reward of state $i$ when following the batting strategy~$\pi$.

\begin{figure} \label{fig:valueiter}\vspace*{-20pt}
\begin{pseudocode}[ruled]{Value Iteration}{P^{\textsc
{\scriptsize Training}},g,\mathcal{S}, U}
\REPEAT
\Delta= 0 \\
\FOREACH i \in\mathcal{S}\cap\mathcal{E}^c \DO\\
\hspace*{8pt} v \gets J(i) \\
\hspace*{8pt} J(i) \gets\max_u \sum_{j} P_u^{\textsc{\scriptsize Training}}(i,j)[g(i,u,j)
+ J(j)] \\
\hspace*{8pt} \Delta\gets\max(\Delta, |v - J(i)|) \\
\UNTIL(\Delta< \varepsilon) \\
\OUTPUT{\mathrm{deterministic\ policy}\   \pi= \{u^0, \ldots,
u^{n-1}\} }
\end{pseudocode}\vspace*{-20pt}
\caption{The Value Iteration algorithm outputs the
batting strategy $\pi$ that maximizes the expectation at each state
over the input data. Reprinted with permission from Richard S. Sutton
and Andrew G. Barto, Reinforcement Learning: An Introduction, published
by The MIT Press.}
\label{fig:valueiteration}
\end{figure}

\begin{figure}[b]\vspace*{-20pt}
\begin{pseudocode}[ruled]{Policy Evaluation}{P^{\textsc
{\scriptsize Test}},g,\mathcal{S}, \pi}
\REPEAT
\Delta= 0 \\
\FOREACH i \in\mathcal{S}\cap\mathcal{E}^c \DO\\
\hspace*{8pt}v \gets J(i) \\
\hspace*{8pt}J(i) \gets \sum_{j} P_{\pi
(i)}^{\textsc{\scriptsize Test}}(i,j) [g(i,\pi(i),j) + J(j)] \\
\hspace*{8pt}\Delta\gets\max(\Delta, |v - J(i)|) \\
\UNTIL(\Delta< \varepsilon) \\
\OUTPUT{\mathrm{J}^\pi}
\end{pseudocode}\vspace*{-20pt}
\caption{The Policy Evaluation algorithm, that calculates
the expected rewards of every state when using the batting strategy $\pi
$, given by $J^\pi\in\mathbb{R}^n$, on the input data. Reprinted with
permission from Richard S. Sutton and Andrew G. Barto, Reinforcement
Learning: An Introduction, published by The MIT Press.}
\label{fig:pealg}
\end{figure}

The transition probability matrices estimated via the training and
test data, represented by $P_u^{\textsc{\scriptsize Training}}(i,j)$ and
$P_u^{\textsc{\scriptsize Test}}(i,j)$ in equations (\ref{eqn:opvalstate}) and
(\ref
{eqn:peeq}), respectively, are the transition probabilities used by Value
Iteration and Policy Evaluation to compute and evaluate, respectively,
the best batting strategy, $\pi$. The batting strategies are estimated
against probability transitions matrices computed from either a
population of at-bats for one season of a single pitcher's data
(pitcher-specific batting strategy) or one season of a population of
pitchers' data (general batting strategy; see Section \ref{sect:stratmeth}).

Policy Evaluation therefore allows a quantitative comparison of
the pitcher-specific and general batting strategies, $\pi^p$ and $\pi
^g$, on the same transition probabilities representing the pitcher's
decision-making. Equation (\ref{eqn:peeq}) shows that the $\{0,0\}$
state's expected reward, denoted by $J(\{0,0\})$, requires the expected
reward of every other state to be calculated first. It follows that
$J(\{0,0\})$ is the expected reward of the entire at-bat. Thus, a
pitcher is exploited if and only if $J^{\pi^p}(\{0,0\}) \geq J^{\pi
^g}(\{0,0\})$ [\citet{sb}] (see Section \ref{sect:comps}).

\subsection{Models}\label{sect:models}
The generality of the Markov process allows us to propose models with
various degrees of resolution, which depends on the number of phenomena
considered at each state. Below we outline two models of information
resolution as useful starting points.
\subsubsection{Simple RLIB (SRLIB)} \label{desc1}

Initially, we represented a pitcher's decisions by conditioning the
pitch outcome (reflected by the future pitch count) on the batter's
action at the previous pitch count. SRLIB therefore has $n = |\mathcal
{S}_S\cap\mathcal{E}^c| = 12$ nonterminal states
\begin{eqnarray*}
\mathcal{S}_S &=& \bigl\{\{0,0\},\{1,0\},\{2,0\},\{3,0\},\{
0,1\},\{0,2\},\{1,1\},\{1,2\},\{2,1\},\{2,2\},\\
&&\hspace*{157pt}\{3,1\},\{3,2\},\mathrm
{O},\mathrm{S},\mathrm{D},\mathrm{T},\mathrm{HR},\mathrm{W} \bigr\}.
\end{eqnarray*}


\subsubsection{Complex RLIB (CRLIB)}\label{desc2}
CRLIB conditions the pitcher's selected pitch-type at the current
pitch count on both the pitch-type and batting actions at the previous
pitch count.

We observed that the MLB GameDay system gave as many as 8
pitch-types for one pitcher. We therefore generalized pitch-types to
four categories: fastball-type, curveball/changeup, sinking/sliding,
and knuckleball/unknown pitches. 
Assuming the set of pitch-types is $\mathcal{T}$, our abstraction
admits at most four pitch-types for every pitcher---that is, $|\mathcal
{T}| \leq4$.

The inclusion of pitch-types results in state space $\mathcal{S}_C
= \{\mathcal{S}_S \cap\mathcal{E}^c\} \times\mathcal{T} \cup\mathcal
{E}$, where every nonterminal state incorporates the four pitch-types
at each pitch count, giving $n = |\{\mathcal{S}_S\cap\mathcal{E}^c\}
\times\mathcal{T}| = 48$ nonterminal states, 
%
\begin{eqnarray*}
\mathcal{S}_C& =&  \bigl\{\{0,0\},\{1,0\},\{2,0\},\{3,0\},\{
0,1\},\{0,2\},\{1,1\},\{1,2\},\{2,1\},\{2,2\},\\
&&\hspace*{243pt} \{3,1\},\{3,2\} \bigr\}\\
&&{}
\times\mathcal{T} \cup\mathcal{E}.
\end{eqnarray*}

We represent the expected reward of the $\{0,0\}$ state as the weighted
average over the expected rewards of the four pitch types associated
with the $\{0,0\}$ state:
\[
J^{\pi}\bigl(\{0,0\}\bigr) = \frac{1}{K}\sum
_{t_i \in\mathcal{T}} K_{t_i}J^{\pi
}\bigl(\{0,0\}\times
t_i\bigr),
\]
where $K_{t_i}$ is the number of times pitch-type $t_i$ was
thrown in the test data and $K$ is the total number of pitches in the
test data.
%

\begin{figure}
{\fontsize{9.5}{11.8}\selectfont
\begin{tabular*}{\textwidth}{@{\extracolsep{\fill}}lc@{}}
$
\pi_{\mathrm{SRLIB}}^p = \left[ \matrix{
\{0,0\}
\vspace*{2pt}\cr
\{1,0\}
\vspace*{2pt}\cr
\{2,0\}
\vspace*{2pt}\cr
\{3,0\}
\vspace*{2pt}\cr
\{0,1\}
\vspace*{2pt}\cr
\{0,2\}
\vspace*{2pt}\cr
\{1,1\}
\vspace*{2pt}\cr
\{1,2\}
\vspace*{2pt}\cr
\{2,1\}
\vspace*{2pt}\cr
\{2,2\}
\vspace*{2pt}\cr
\{3,1\}
\vspace*{2pt}\cr
\{3,2\}} \right] = \left[ \matrix{
0
\vspace*{2pt}\cr
1
\vspace*{2pt}\cr
0
\vspace*{2pt}\cr
0
\vspace*{2pt}\cr
0
\vspace*{2pt}\cr
0
\vspace*{2pt}\cr
0
\vspace*{2pt}\cr
0
\vspace*{2pt}\cr
1
\vspace*{2pt}\cr
0
\vspace*{2pt}\cr
0
\vspace*{2pt}\cr
0}\right]
$&
$
\pi_{\mathrm{CRLIB}}^p = \left[ \matrix{
\left[ \matrix{ 0 & 1 & 0 & 0 & 0 & 1 & 0 & 0 & 0
& 1 & 0 & 0 }\right ]^{\top}
\vspace*{2pt}\cr
[ \mathrm{null} ]^{\top}
\vspace*{2pt}\cr
\left[ \matrix{ 0 & 0 & 1 & 1 & 1 & 0 & 0 & 0 & 1
& 0 & 1 & 0 } \right]^{\top}
\vspace*{2pt}\cr
[ \mathrm{null} ]^{\top}
} \right]
$\\\\
$
\pi_{\mathrm{SRLIB}}^g = \left[\matrix{
\{0,0\}
\vspace*{2pt}\cr
\{1,0\}
\vspace*{2pt}\cr
\{2,0\}
\vspace*{2pt}\cr
\{3,0\}
\vspace*{2pt}\cr
\{0,1\}
\vspace*{2pt}\cr
\{0,2\}
\vspace*{2pt}\cr
\{1,1\}
\vspace*{2pt}\cr
\{1,2\}
\vspace*{2pt}\cr
\{2,1\}
\vspace*{2pt}\cr
\{2,2\}
\vspace*{2pt}\cr
\{3,1\}
\vspace*{2pt}\cr
\{3,2\} } \right] = \left[ \matrix{ %
0
\vspace*{2pt}\cr
0
\vspace*{2pt}\cr
1
\vspace*{2pt}\cr
1
\vspace*{2pt}\cr
0
\vspace*{2pt}\cr
0
\vspace*{2pt}\cr
0
\vspace*{2pt}\cr
0
\vspace*{2pt}\cr
0
\vspace*{2pt}\cr
0
\vspace*{2pt}\cr
0
\vspace*{2pt}\cr
0} \right]
$&
$
\pi_{\mathrm{CRLIB}}^g = \left[ \matrix{ %
\left[ \matrix{ 0 & 0 & 1 & 0 & 1 & 0 & 1 & 0 & 1
& 0 & 1 & 1 }\right ]^{\top}
\vspace*{2pt}\cr
\left[ \matrix{ 0 & 0 & 0 & 1 & 0 & 0 & 1 & 1 & 1
& 0 & 1 & 1 }\right ]^{\top}
\vspace*{2pt}\cr
\left[ \matrix{ 0 & 0 & 0 & 0 & 0 & 0 & 1 & 0 & 1
& 1 & 1 & 1 }\right ]^{\top}
\vspace*{2pt}\cr
[ \mathrm{null} ]^{\top}} \right]
$
\end{tabular*}}
\caption{The pitcher-specific and general strategies,
denoted by $\pi^p$ and $\pi^g$, respectively, computed on SRLIB and
CRLIB for the 2009 season. CRLIB's pitcher-specific batting strategy
for Roy Halladay had an empty second and fourth row because the MLB
GameDay system reported he did not throw any Sinking/sliding or
Knuckleball/unknown type pitches in the training data; a similar
argument holds for the general batting strategy's fourth row.}
\label{fig:policies}
\end{figure}

\subsection{Illustration of batting strategies}\label{illust}
We now illustrate the batting strategies that are produced by
either SRLIB or CRLIB, respectively. As shown in Figure \ref{fig:policies}, SRLIB and CRLIB have \textit{n}-dimensional and ($n
\times4$)-dimensional batting strategies, respectively. The action at
each state is represented by a binary value corresponding to Stand (0)
and Swing (1).

For the strategies given in Figure \ref{fig:policies},
CRLIB/SRLIB's pitcher-specific batting strategy exploited Roy Halladay
on the 2010 test data. However, only CRLIB exploits Halladay on the
2008 test data, presumably because it incorporates information about
his pitch selection, as it is believed that pitchers often rely on
their ``best pitches'' in specific pitch counts. For example: If
Halladay throws a fastball in the $\{2,2\}$ pitch count, $\pi^p_{\mathrm{SRLIB}}$
selects the batting action $u=\operatorname{Stand}$, whereas $\pi^p_{\mathrm{CRLIB}}$
selects $u=\operatorname{Swing}$. In comparison to the SRLIB model, we see
that the inclusion of pitch type gives the CRLIB model an enriched
representation that can improve a batter's opportunity of reaching base.

\subsection{Comparing an ``intuitive'' and general batting
strategy}\label{sect:comps}

We show that the general batting strategy is a more competitive
baseline performance measure than an intuitive batting strategy. This
intuitive batting strategy selects the action Swing at states $\{1,0\},\{
2,0\},\{3,0\},\{2,1\},\{3,1\}$, and Stand in states $\{0,0\},\{1,1\},\{
1,2\}, \{2,2\},\{3,2\},\{0,1\},\{0,2\}$. In other words, the intuitive
batting strategy reflects the intuition that the batter should only
swing in a batter's count;\footnote{Please see \hyperref[sect:lingo]{Appendix} for baseball terminology.}
we show that these types of batting
strategies are inferior to statistically computed ones, such as the
general batting strategy.

After performing Policy Evaluation for both batting strategies, we
observed that the general batting strategy outperformed the intuitive
batting strategy 146 out of 150 times. The general batting strategy's
dominant performance justifies its selection as the competitive
baseline performance measure. Thus, we cannot assume that the
pitcher-specific batting strategy will perform better than, or even
equal to, the general batting strategy when performing Policy
Evaluation on the respective pitcher's test data. 

Since each pitcher's ``best pitch(es)'' can vary, it follows that
the probability distributions over future states from the current state
will also vary, especially in states with a batter's count. This is a
consequence of the uniqueness of each pitcher's behaviour, which is
reflected through their pitch selection at each state in the at-bat,
and thereby quantified in their respective transition probabilities.
This is personified by R.~A. Dickey, as he only has one ``best pitch''
(knuckleball) and throws other pitch types to make his behaviour less
predictable. Thus, when Dickey's knuckleball is ineffective, an optimal
policy will recommend swinging at every nonknuckle pitch, as it is
computed over data that shows an improved outcome for the batter. In
contrast, pitchers that consistently throw more than one pitch type for
strikes, such as Roy Halladay, are more difficult to exploit via the
pitcher-specific batting strategy because the improved outcome in
swinging at their ``weaker'' pitches is marginal in comparison to
pitchers that consistently throw fewer pitch types for strikes. Given
that the pitcher-specific and general batting strategies are computed
over different populations,\footnote{Section \ref{sect:intro}, second
last paragraph of page 2.} and each at-bat contains information about
the respective pitcher's behaviour, it follows that pitchers are not
equally susceptible to being exploited in an empirical setting. 

If the pitcher-specific batting strategy performs equal to the
general batting strategy, the competitiveness/dominance of the latter
over intuitively-constructed batting strategies ensures that the
respective pitcher is still \textit{exploited} because the intuitive
batting strategy can be viewed as the standard strategy that is
employed in baseball. Let $\pi^i$ be the intuitive batting strategy and
assume that the pitcher-specific batting strategy performs equal to the
general batting strategy. By transitivity, $J^{\pi^g}(\{0,0\}) \gg
J^{\pi^i}(\{0,0\})$, and $J^{\pi^p}(\{0,0\}) == J^{\pi^g}(\{0,0\})$,
then $J^{\pi^p}(\{0,0\}) \gg J^{\pi^i}(\{0,0\})$. 
%


\section{Spatial component}\label{spatdesc}

\begin{figure}[b]

\includegraphics{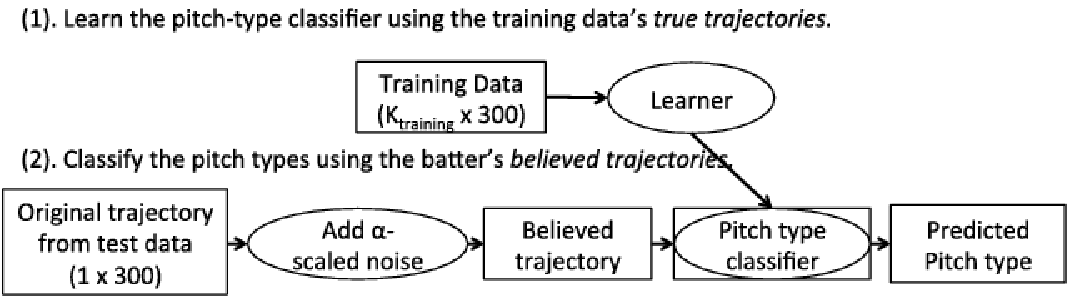}

\caption{Illustrating our spatial model, where $\alpha$ is
parameterized according to the batter, $K_{\mathrm{training}}$ denotes the
number of pitches in the respective pitcher's training data, and 300
represents the $m=100$ three-dimensional points that describe
the pitch trajectory. Here, the Learner is a quadratic kernel Support
Vector Machine (SVM).}
\label{fig:spatmodel}
\end{figure}

A spatial component is introduced to highlight the utility of the
exploited pitchers' pitcher-specific batting strategies. The spatial
component associates the pitch-type based on the respective pitch's
spatial trajectory, where this trajectory is parameterized specifically
to the batter being simulated. Thus, the spatial component predicts the
pitch-type for the batter-parameterized spatial trajectory given as
input. This allows us to simulate a batter's performance when they use
the exploited pitcher's pitcher-specific batting strategy against the
respective pitcher (see Figure~\ref{fig:spatmodel}).

The spatial information for each pitch contains the
three-dimensional acceleration, velocity, starting and ending
positions. This information is obtained from the MLB GameDay system
after it fits a quadratic polynomial to 27 instantaneous images
representing the pitch's spatial trajectory. These pictures are taken
by cameras on opposite sides of the field. The MLB GameDay system
therefore performs a quadratic fit to the trajectory data. 

\begin{figure}

\includegraphics{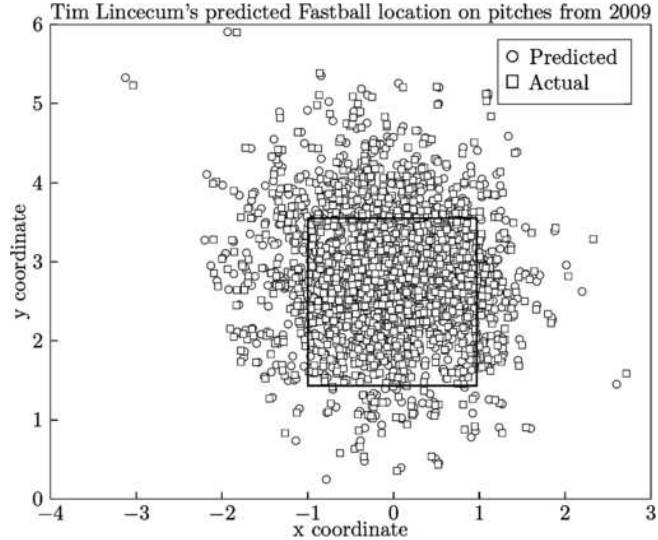}

\caption{The actual (square) and predicted (circle)
locations for Tim Lincecum's pitches in the 2009 season, when
performing OLS regression on break angle, initial velocity, and break
length features to predict the end location of each pitch. The
rectangle in the center of the plot represents the strike zone.}\label
{fig:olspred}
\end{figure}

There are two problems with this fit. First, acceleration is
assumed to be constant, which is certainly not true. Second, there
exists a near-perfect correlation between variables obtained from the
fit (such as velocity and the end location of a pitch) and the
independent variable (acceleration) used to produce this fit.
Figure \ref{fig:olspred} displays pitch locations along with predicted
values from regression with velocity or acceleration as predictors and
location as the response.

We see that the quadratic fit severely limits the use of each
pitch's spatial trajectory. To address this limitation, we use the
instantaneous positions of every pitch trajectory to predict the pitch type.

\subsection{Batter-parameterized pitch identification}

We add $\alpha$-scaled noise, where $\alpha$ is a batter-specific
parameter, to the original spatial trajectory of each pitch to
represent the respective batter's believed trajectories. This is
achieved by first drawing independent, identically distributed noise
from the uniform distribution on the $[-1,1]$ interval, which is then
multiplied by the parameter $\alpha$, and finally added to the \textit
{m} evenly-spaced, three-dimensional, instantaneous positions of the
original (true) pitch trajectory.

For the batter's believed trajectories to accurately represent
their pitch-identification ability, $\alpha$ is defined as the number
of strikeouts divided by the number of plate appearances on the same
year which the batting strategy is computed on. Only considering
strikeouts is justified by the fact that any recorded out from putting
the ball in play implies that a player identified the ball accurately
enough to achieve contact. In contrast, a batter that strikes out
either failed to identify the pitch as a strike or failed to establish
contact with the ball. Thus, adding $\alpha$-scaled noise to original
trajectory reproduces a batter's believed trajectory.

The training data's 
true trajectories
are given as input to a Support Vector Machine (SVM) [\citet{vapnik},
Hastie, Tibshirani and Friedman (\citeyear{htf})]
with a quadratic kernel, where each trajectory's $m=100$
three-dimensional instantaneous positions form a single data point that
has an associated pitch-type. The SVM algorithm then computes a spatial
classifier, which is composed of coefficients $\bolds\beta=
[\beta_0,\ldots,\beta_{300}]$, that best separates the training data's
original trajectories according to pitch-type, 
usually with high
accuracy.\footnote{Training accuracies are provided in Table \ref{fig:spatrez}.}

%
\begin{sidewaystable}
\tablewidth=\textwidth
\caption{The annual statistics from the 2008, 2009 and
2010 seasons for the 25 pitchers used in our evaluation\protect\tabnoteref[*]{ttl1}}
\label{fig:stats}
\begin{tabular*}{\textwidth}{@{\extracolsep{\fill}}ld{1.2}d{1.3}d{2.0}d{2.0}d{3.1}d{1.2}d{1.3}d{2.0}d{2.0}d{3.1}d{1.2}d{1.3}d{2.0}d{2.0}d{3.1}@{}}
\hline
 & \multicolumn{5}{c}{\textbf{2008}} & \multicolumn{5}{c}{\textbf{2009}} &
 \multicolumn{5}{c@{}}{\textbf{2010}} \\[-6pt]
 & \multicolumn{5}{c}{\hrulefill} & \multicolumn{5}{c}{\hrulefill} &
 \multicolumn{5}{c@{}}{\hrulefill} \\
\multicolumn{1}{@{}l}{\textbf{Player}} &
\multicolumn{1}{c}{\textbf{ERA}} & \multicolumn{1}{c}{\textbf{WHIP}} &
\multicolumn{1}{c}{\textbf{W}} &
\multicolumn{1}{c}{\textbf{L}} & \multicolumn{1}{c}{\textbf{IP}} &
\multicolumn{1}{c}{\textbf{ERA}} &
\multicolumn{1}{c}{\textbf{WHIP}} & \multicolumn{1}{c}{\textbf{W}} &
\multicolumn{1}{c}{\textbf{L}} &
\multicolumn{1}{c}{\textbf{IP}} &
\multicolumn{1}{c}{\textbf{ERA}} & \multicolumn{1}{c}{\textbf{WHIP}} &
\multicolumn{1}{c}{\textbf{W}} & \multicolumn{1}{c}{\textbf{L}} &
\multicolumn{1}{c@{}}{\textbf{IP}} \\
\hline
Roy Halladay & 2.78 & 1.053 & 20 & 11 & 246.0 & 2.79 & 1.126 & 17 & 10
& 239.0 & 2.44 & 1.041 & 21 & 10 & 250.2 \\
Cliff Lee & 2.54 & 1.110 & 22 & 3 & 223.1 & 3.22 & 1.243 & 14 & 13 &
231.2 & 3.18 & 1.003 & 12 & 9 & 212.1 \\
Cole Hamels & 3.09 & 1.082 & 14 & 10 & 227.1 & 4.32 & 1.286 & 10 & 11
& 193.2 & 3.06 & 1.179 & 12 & 11 & 208.2 \\
Jon Lester & 3.21 & 1.274 & 16 & 6 & 210.1 & 3.41 & 1.230 & 15 & 8 &
203.1 & 3.25 & 1.202 & 19 & 9 & 208.0 \\
Zack Greinke & 3.47 & 1.275 & 13 & 10 & 202.1 & 2.16 & 1.073 & 16 & 8
& 229.1 & 4.17 & 1.245 & 10 & 14 & 220.0 \\
Tim Lincecum & 2.62 & 1.172 & 18 & 5 & 227.0 & 2.48 & 1.047 & 15 & 7 &
225.1 & 3.43 & 1.272 & 16 & 10 & 212.1 \\
CC Sabathia & 2.70 & 1.115 & 17 & 10 & 253.0 & 3.37 & 1.148 & 19 & 8 &
230.0 & 3.18 & 1.191 & 21 & 7 & 237.2 \\
Johan Santana & 2.53 & 1.148 & 16 & 7 & 234.1 & 3.13 & 1.212 & 13 & 9
& 166.2 & 2.98 & 1.170 & 11 & 9 & 199.0 \\
Felix Hernandez & 3.45 & 1.385 & 9 & 11 & 200.2 & 2.49 & 1.135 & 19 &
5 & 238.8 & 2.27 & 1.057 & 13 & 12 & 249.2 \\
Chad Billingsley & 3.14 & 1.336 & 16 & 10 & 200.2 & 4.03 & 1.319 & 12
& 11 & 196.1 & 3.57 & 1.278 & 12 & 11 & 191.2 \\
Jered Weaver & 4.33 & 1.285 & 11 & 10 & 176.2 & 3.75 & 1.242 & 16 & 8
& 211.0 & 3.01 & 1.074 & 13 & 12 & 224.1 \\
Clayton Kershaw & 4.26 & 1.495 & 5 & 5 & 107.2 & 2.79 & 1.229 & 8 & 8
& 171.0 & 2.91 & 1.179 & 13 & 10 & 204.1 \\
Chris Carpenter & 1.76 & 1.304 & 0 & 1 & 15.1 & 2.24 & 1.007 & 17 & 4
& 192.2 & 3.22 & 1.179 & 16 & 9 & 235.0 \\
Matt Garza & 3.70 & 1.240 & 11 & 9 & 184.2 & 3.95 & 1.261 & 8 & 12 &
203.0 & 3.91 & 1.251 & 15 & 10 & 204.2 \\
Adam Wainwright & 3.20 & 1.182 & 11 & 3 & 132.0 & 2.63 & 1.210 & 19 &
8 & 233.0 & 2.42 & 1.051 & 20 & 11 & 230.1 \\
Ubaldo Jimenez & 3.99 & 1.435 & 12 & 12 & 198.2 & 3.47 & 1.229 & 15 &
12 & 218.0 & 2.88 & 1.155 & 19 & 8 & 221.2 \\
Matt Cain & 3.76 & 1.364 & 8 & 14 & 217.2 & 2.89 & 1.181 & 14 & 8 &
217.2 & 3.14 & 1.084 & 13 & 11 & 223.1 \\
Jonathan Sanchez & 5.01 & 1.449 & 9 & 12 & 158.0 & 4.24 & 1.365 & 8 &
12 & 163.1 & 3.07 & 1.231 & 13 & 9 & 193.1 \\
Roy Oswalt & 3.54 & 1.179 & 17 & 10 & 208.2 & 4.12 & 1.241 & 8 & 6 &
181.1 & 2.76 & 1.025 & 13 & 13 & 211.2 \\
Justin Verlander & 4.84 & 1.403 & 11 & 17 & 201.0 & 3.45 & 1.175 & 19
& 9 & 240.0 & 3.37 & 1.163 & 18 & 9 & 224.1 \\
Josh Johnson & 3.61 & 1.351 & 7 & 1 & 87.1 & 3.23 & 1.158 & 15 & 5 &
209.0 & 2.30 & 1.105 & 11 & 6 & 183.2 \\
John Danks & 3.32 & 1.226 & 12 & 9 & 195.0 & 3.77 & 1.283 & 13 & 11 &
200.1 & 3.72 & 1.216 & 15 & 11 & 213.0 \\
Edwin Jackson & 4.42 & 1.505 & 14 & 11 & 183.1 & 3.62 & 1.262 & 13 & 9
& 214.0 & 4.47 & 1.395 & 10 & 12 & 209.1 \\
Max Scherzer & 3.05 & 1.232 & 0 & 4 & 56.0 & 4.12 & 1.344 & 9 & 11 &
170.1 & 3.50 & 1.247 & 12 & 11 & 195.2 \\
Ted Lilly & 4.09 & 1.226 & 17 & 9 & 204.2 & 3.10 & 1.056 & 12 & 9 &
177.0 & 3.62 & 1.079 & 10 & 12 & 193.2 \\
\hline
\end{tabular*}
\tabnotetext[*]{ttl1}{Please see \hyperref[sect:lingo]{Appendix} for
baseball terminology.}
\end{sidewaystable}

The spatial classifier allows pitch-type identification to be
standardized across batters, because the respective batter's believed
trajectory is only identified as the correct pitch-type if it is
similar to the original trajectory. If the believed trajectory differs
enough from the original trajectory, it will be identified as the
incorrect pitch-type; this is reflective of players with higher $\alpha
$ values that strike out often. We can therefore view the spatial
classifier as an oracle.

\section{Methods and evaluation}\label{methods}

\subsection{Strategic component}\label{sect:stratmeth}
For our evaluation, we use 3 years of pitch-by-pitch data for 25
elite\footnote{We remind readers that our definition of elite is
heuristically defined.} pitchers, as shown in Table \ref{fig:stats}. We
evaluated the batting strategies performance for all six unique
combinations of the training and test data, which gave a total of
$25\times6 = 150$ pitcher-specific batting strategies. The data was
obtained from MLB's GameDay system, which provided three complete
seasons of pitching data, containing the pitch outcome, pitch-type,
number of balls and strikes, and the batter's actions. The technical
details of the data collection and formatting process, which is
necessary before applying the Reinforcement Learning algorithms, are
provided in the supplement [\citet{suppref}].

Initially the general batting strategy was trained over all
pitchers data for the respective season, where it was observed that the
general batting strategy outperformed the pitcher-specific batting
strategies on the respective season by a significant margin. In other
words, training the general batting strategy over all of the pitchers'
annual data led to invalidation of the hypothesis, regardless of
whether pitchers were equally exploitable. Given that the general
batting strategy was trained over a much larger data set than the
pitcher-specific batting strategy, we sought a way to standardize the
training data used to compute the general batting strategy, which is
described in the following paragraph. We acknowledge that in real-world
settings, each of the batting strategies should be trained over \textit
{all} available information, as this information improves the
resolution and precision of the probability estimates exploited by the
batting strategy.

The size of the general batting strategy's training data is
approximately equal to the average number of pitches thrown by the 25
pitchers in the respective season, where every pitcher's at-bats in
this data set were randomly sampled from all of their at-bats for the
respective season. To ensure the general batting strategy's training
data is representative of all pitchers' data for the respective season,
sampling terminates after adding an at-bat for which the total number
of pitches exceeds the proportion of pitches that should be contributed
by the pitcher. For example, Roy Halladay threw 3319 pitches in 2009,
and all 25 pitchers threw a total of 80,879 pitches. Roy Halladay
therefore contributes $\lceil{0.04103\times3235.16}\rceil= 133$
pitches to the data set. Repeating this process for all 25 pitchers
2009 data gives a data set comprised of 3276 pitches.

To address the possibility that the aggregate data sample contains
unrepresentative at-bats for pitcher(s), which would misrepresent the
performance of the general batting strategy, the aggregate data sample
was independently constructed 10 times, where the general batting
strategy was computed against each of the 10 (training) data samples.
Thus, the general batting strategy performance is the average over the
10 general batting strategies' expected rewards on the test data.

We evaluate the hypothesis under two different assumptions: all
pitchers are equally susceptible to being exploited, and all pitchers
are \textit{not} equally susceptible to being exploited, where we
believe the latter assumption is more relevant to the real-world
setting (see Section \ref{sect:comps} for our explanation). The null
hypothesis is the same under either assumption, but the alternative
hypothesis $H_1$ is slightly different:
\begin{longlist}
\item[$H_0$:]$p=\frac{1}{2}$. The pitcher-specific and general
batting strategy are equally likely to exploit the respective pitcher.
\item[$H_1$:]$p>\frac{1}{2}$. The pitcher-specific batting
strategy will perform:
\begin{enumerate}[\quad]
\item[]\textit{strictly} better than the general batting
strategy more than 50\% of the time (assuming that pitchers are \textit
{equally} exploitable).
\item[] better than \textit{or equal to} the general batting strategy
more than 50\% of the time (assuming that pitchers are \textit{not}
equally exploitable).
\end{enumerate}
\end{longlist}
For both hypotheses, the \textit{p}-value calculation is given by
\[
P(X > M) = \sum_{i=M+1}^{150} \pmatrix{150\cr i}
p^i(1-p)^{150-i},
\]
where \textit{M} is the number of pitcher-specific batting
strategies that exploit the respective pitcher(s). Under the assumption
that the pitchers are equally exploitable, $H_0$ could \textit{not} be
rejected; when assuming that pitchers are not equally exploitable,
$H_0$ was rejected \textit{only} for the CRLIB model ($P(X > M) =
3e\mbox{--}2$, $M=87$).

When Policy Evaluation is performed on the pitcher-specific or
general batting strategy, the pitch-type thrown at the current pitch
count is given before selecting the batting action. It follows that
Policy Evaluation implicitly assumes that the pitch-types are always
identified correctly. This was a desirable assumption because it only
considers the strategic aspect of the at-bat when calculating the
expected rewards for the respective strategy. It follows that the
hypothesis evaluation is completely independent of the batters used in
the evaluation.

\subsection{Simulating batting strategies with the spatial
component}\label{sect:stratspat}

We define the chance threshold as the proportion of the majority
pitch-type thrown in the training data, which serves as a baseline for
a batter's pitch-type identification ability. We only simulate batters
whose believed trajectories from the test data are classified with an
accuracy above the chance threshold. This stipulation reflects our
requirement that a batter must be able to identify pitch-types better
than guessing the majority pitch-type. Table \ref{fig:spatrez} provides
the classification accuracies for the batters included in our simulation.

%
\begin{sidewaystable}
\tablewidth=\textwidth
\caption{Evaluating the pitch identification ability on
the 2009 (test) data, after training the spatial classifier on 2010
data. The accuracy is the number of correctly predicted pitch-types on
the respective batter's believed trajectories from the 2009 season,
after only observing the original trajectories from 2010. The chance
threshold is provided to show ``how much better'' the batter is doing
than blindly guessing one class\protect\tabnoteref[*]{ttl2}}\label{fig:spatrez}
\begin{tabular*}{\textwidth}{@{\extracolsep{\fill}}lcccccc@{}}
\hline
&&& \multicolumn{2}{c}{\textbf{Accuracy}} && \\[-6pt]
&&& \multicolumn{2}{c}{\hrulefill} && \\
\multicolumn{1}{@{}l}{\textbf{Batter}} &
\multicolumn{1}{c}{\textbf{Pitcher}} &
\multicolumn{1}{c}{$\bolds{\alpha}$} &
\multicolumn{1}{c}{\textbf{Training}} &
\multicolumn{1}{c}{\textbf{Test}} &
\multicolumn{1}{c}{\textbf{Chance threshold}}&
\multicolumn{1}{c@{}}{\textbf{\# of PA (in 2009)}}\\
\hline
Miguel Cabrera & Zack Greinke & 0.1466 & $2776/2903$ (95.63\%) &
$3135/3376$ (92.86\%) & $2005/3376$ (59.39\%) & 14 \\
Joey Votto & Roy Oswalt & 0.1929 & $3050/3374$ (90.4\%) & $1926/3151$
(61.12\%) & $1787/3151$ (56.71\%) & 12 \\
Joe Mauer & Justin Verlander & 0.0908 & $3545/3633$ (97.58\%) &
$3664/3794$ (96.57\%) & $2553/3794$ (67.29\%) & 14 \\
Ichiro Suzuki & CC Sabathia & 0.1175 & $3154/3736$ (84.42\%) & $3089/3985$
(71.86\%) & $2405/3095$ (54.99\%) & 11 \\
Jose Bautista & Jon Lester & 0.1698 & $3047/3263$ (93.38\%) & $2486/3397$
(73.18\%) & $2423/3397$ (71.33\%) & 11 \\
Derek Jeter & Matt Garza & 0.1434 & $3201/3331$ (96.1\%) & $3125/3288$
(95.04\%) & $2336/3288$ (71.05\%) & 14 \\
Prince Fielder & Matt Cain & 0.1933 & $3265/3782$ (95.85\%) & $3019/3177$
(95.03\%) & $1986/3177$ (62.51\%) & \phantom{0}9 \\
Matt Holliday & Clayton Kershaw & 0.1378 & 3291/3328 (98.89\%) &
2872/3062 (93.79\%) & 2163/3062 (70.64\%) & \phantom{0}8 \\
Ryan Howard & Tim Lincecum & 0.2532 & $3633/3870$ (93.88\%) & $3047/3338$
(91.28\%) & $1874/3338$ (56.14\%) & \phantom{0}7 \\
Mark Teixeira & Cliff Lee & 0.1713 & $3292/3427$ (96.06\%) & $3134/3965$
(79.04\%) & $2542/3965$ (64.11\%) & 14 \\
Nick Markakis & Jon Lester & 0.1311 & $3047/3263$ (93.38\%) & $2475/3397$
(72.86\%) & $2423/3397$ (71.33\%) & 13 \\
Carlos Gonzalez & Matt Cain & 0.2123 & $3265/3782$ (95.85\%) & $3008/3177$
(94.68\%) & $1986/3177$ (62.51\%) & \phantom{0}7 \\
Evan Longoria & Roy Halladay & 0.1876 & $3712/3782$ (98.15\%) &
$3146/3319$ (94.79\%) & $2445/3319$ (73.67\%) & 23 \\
Brandon Philips & Chris Carpenter & 0.1208 & $2884/3420$ (84.33\%) &
$1438/2623$ (54.82\%) & $1227/2623$ (46.78\%) & 11 \\
Manny Ramirez & Jonathan Sanchez & 0.1879 & $3420/3529$ (96.91\%) &
$2391/2724$ (87.78\%) & $1866/2724$ (68.50\%) & 11 \\
Adrian Gonzalez & Matt Cain & 0.1645 & $3265/3782$ (95.85\%) & $3042/3177$
(95.75\%) & $1986/3177$ (62.51\%) & \phantom{0}7 \\
Carl Crawford & CC Sabathia & 0.1569 & $3154/3736$ (84.42\%) & $3061/3985$
(76.81\%) & $2405/3985$ (60.35\%) & 13 \\
Troy Tulowitzki & Clayton Kershaw & 0.1475 & $3291/3328$ (98.89\%) &
$2883/3062$ (94.15\%) & $2163/3062$ (70.64\%) & 14 \\
Matt Kemp & Jonathan Sanchez & 0.2545 & $3420/3529$ (96.91\%) &
$2394/2724$ (87.89\%) & $1866/2724$ (68.50\%) & 11 \\
Alex Rodriguez & Roy Halladay & 0.1647 & $3712/3782$ (98.15\%) &
$3147/3319$ (94.82\%) & $2445/3319$ (73.67\%) & 18 \\
\hline
\end{tabular*}
\tabnotetext[*]{ttl2}{Please see \hyperref[sect:lingo]{Appendix} for baseball terminology.}
\end{sidewaystable}

\begin{figure}

\includegraphics{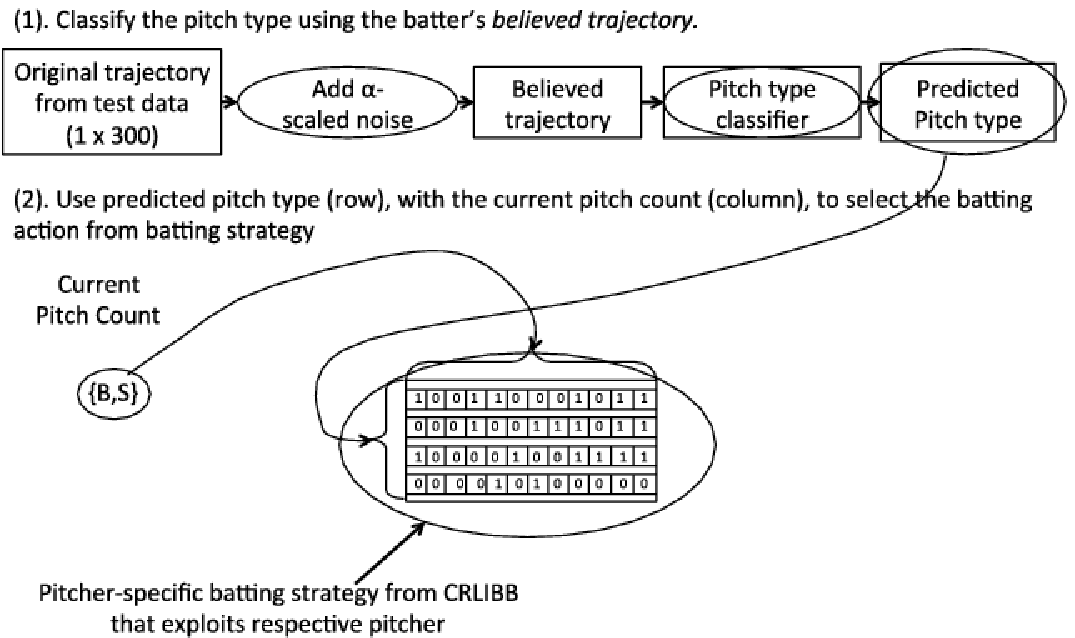}

\caption{Illustrating the simulation for a batter using an exploited
pitcher's pitcher-specific batting strategy. The spatial classifier's
predicted pitch-type is used to select the row that contains the
batting actions for the respective pitch-type, and the current pitch
count is used to select the batting action for this pitch-type.}
\label{fig:both}
\end{figure}

We use the spatial component to evaluate twenty prominent batters.
Among these batters, ten are considered elite and ten are considered to
be nonelite. We simulate each batter's performance against an exploited
pitcher when using the respective pitcher-specific batting strategy.
For every at-bat that is simulated, the spatial component first
predicts the pitch-type using the respective batter's believed
trajectory. Then, this predicted pitch-type is used with the current
state to select the appropriate action from the pitcher-specific
batting strategy; see Figure~\ref{fig:both}.

After identifying the pitch-type and selecting the batting action,
the conditional distribution over the future states in the at-bat
becomes determined. We assign the future states ``bin'' lengths that are
equal to the probability of transitioning to their respective states,
where each bin is a disjoint subinterval on [0,1]. We then generate a
random value from the uniform distribution on the [0,1] interval and
select the future state whose subinterval contains this random value.
To ensure that the performance of the batter versus pitcher is
representative of each state's distribution over future states, every
at-bat is simulated 100 times.

\section{Results}\label{results}

\subsection{Strategic component}
The number of pitcher-specific batting strategies that exploited
the respective pitcher are given in Table \ref{fig:ranks}. The raw
results for SRLIB/CRLIB are given in Tables \ref{fig:sRLIBres} and \ref
{fig:cRLIBres}, respectively.

%
\begin{table}[b]
\caption{The number of exploited pitchers using the
respective train/test data partitions}\label{fig:ranks}
\begin{tabular*}{\textwidth}{@{\extracolsep{\fill}}lcccccc@{}}
\hline
& \multicolumn{6}{c@{}}{\textbf{Year of computed batting strategy}} \\[-6pt]
& \multicolumn{6}{c@{}}{\hrulefill} \\
\multicolumn{1}{@{}l}{\textbf{Model}} & \multicolumn{2}{c}{\textbf{2008}} & \multicolumn
{2}{c}{\textbf{2009}} & \multicolumn{2}{c}{\textbf{2010}} \\
\hline
 & 2009 & 2010 & 2008 & 2010 & 2008 & 2009 \\
SRLIB & 11 & 10 & 5 & 12 & 11 & 14 \\
CRLIB & 12 & 15 & 11 & 16 & 16 & 16 \\
\hline
\end{tabular*}
\end{table}

\begin{sidewaystable}
\tablewidth=\textwidth
\caption{Values obtained from SRLIB
after training on one year of data and testing on another. Superscripts
$\pi^p, \pi^g$ denote the expected
reward from the $\{0,0\}$ state when following the pitcher-specific or
general batting strategy. The bolded
pitcher-specific batting strategies are batting strategies that exploit
the respective pitcher on
the respective training and test dataset pair}\label{fig:sRLIBres}
{\fontsize{8.3}{10.3}{\selectfont
\begin{tabular*}{\textwidth}{@{\extracolsep{\fill}}lcccccccccccc@{}}
\hline
& \multicolumn{4}{c}{\textbf{Training on 2008 data}} &
\multicolumn{4}{c}{\textbf{Training on 2009 data}} &
\multicolumn{4}{c@{}}{\textbf{Training on 2010 data}} \\[-6pt]
& \multicolumn{4}{c}{\hrulefill} &
\multicolumn{4}{c}{\hrulefill} &
\multicolumn{4}{c@{}}{\hrulefill} \\
 & \multicolumn{2}{c}{\textbf{2009}} & \multicolumn{2}{c}{\textbf{2010}} &
\multicolumn{2}{c}{\textbf{2008}} & \multicolumn{2}{c}{\textbf{2010}} & \multicolumn
{2}{c}{\textbf{2008}} & \multicolumn{2}{c}{\textbf{2009}} \\[-6pt]
 & \multicolumn{2}{c}{\hrulefill} & \multicolumn{2}{c}{\hrulefill} &
\multicolumn{2}{c}{\hrulefill} & \multicolumn{2}{c}{\hrulefill} & \multicolumn
{2}{c}{\hrulefill} & \multicolumn{2}{c@{}}{\hrulefill} \\
\textbf{Player}& \multicolumn{1}{c}{$\bolds{J^{\pi^g}}$} &
\multicolumn{1}{c}{$\bolds{J^{\pi^p}}$} &
\multicolumn{1}{c}{$\bolds{J^{\pi^g}}$} &
\multicolumn{1}{c}{$\bolds{J^{\pi^p}}$} &
\multicolumn{1}{c}{$\bolds{J^{\pi^g}}$} &
\multicolumn{1}{c}{$\bolds{J^{\pi^p}}$} &
\multicolumn{1}{c}{$\bolds{J^{\pi^g}}$} &
\multicolumn{1}{c}{$\bolds{J^{\pi^p}}$}
&\multicolumn{1}{c}{$\bolds{J^{\pi^g}}$} &
\multicolumn{1}{c}{$\bolds{J^{\pi^p}}$} &
\multicolumn{1}{c}{$\bolds{J^{\pi^g}}$} &
\multicolumn{1}{c}{$\bolds{J^{\pi^p}}$}\\
\hline
Roy Halladay & 0.655 & 0.604 & 0.546 & 0.491 & 0.659 & 0.569 & 0.569 &
$\mathbf{0.766}$& 0.699 & 0.560 & 0.638 & $\mathbf{0.707}$\\
Cliff Lee & 0.692 & $\mathbf{0.706}$ & 0.566 & $\mathbf{0.622}$ & 0.680
& 0.622 & 0.558 & 0.499& 0.678 & 0.664 & 0.707 & 0.571\\
Cole Hamels & 0.716 & $\mathbf{0.768}$ & 0.814 & 0.795 & 0.731 & 0.700
& 0.807 & 0.764& 0.724 & 0.719 & 0.700 & 0.680\\
Jon Lester & 0.713 & $\mathbf{0.764}$ & 0.876 & 0.847 & 0.742 & 0.737 &
0.831 & 0.718& 0.739 & $\mathbf{0.741}$ & 0.710 & 0.709\\
Zack Greinke & 0.642 & 0.608 & 0.747 & $\mathbf{0.835}$ & 0.795 & 0.688
& 0.739 & 0.719& 0.824 & $\mathbf{0.836}$ & 0.607 &$\mathbf{0.672}$\\
Tim Lincecum & 0.649 & $\mathbf{0.694}$ & 0.822 & 0.785 & 0.717 & 0.694
& 0.778 & 0.723& 0.686 & 0.589 & 0.653 & 0.565\\
CC Sabathia & 0.810 & 0.718 & 0.802 & 0.698 & 0.657 & 0.550 & 0.792 &
$\mathbf{0.821}$& 0.593 & 0.550 & 0.825 & $\mathbf{0.836}$\\
Johan Santana & 0.785 & $\mathbf{0.862}$ & 0.773 & 0.703 & 0.680 &
$\mathbf{0.694}$ & 0.774 & $\mathbf{0.838}$ & 0.651 & $\mathbf{0.773}$
& 0.744 & $\mathbf{0.852}$\\
Felix Hernandez & 0.721 & $\mathbf{0.784}$ & 0.823 & 0.786 & 0.777 &
0.765 & 0.815 & 0.786 & 0.717 & 0.662 & 0.760 & 0.613\\
Chad Billingsley & 0.802 & $\mathbf{0.865}$ & 0.788 & $\mathbf{0.812}$
& 0.750 & $\mathbf{0.790}$ & 0.792 & $\mathbf{0.812}$ & 0.752 & $\mathbf
{0.785}$ & 0.792 & $\mathbf{0.860}$\\
Jered Weaver & 0.860 & 0.841 & 0.691 & $\mathbf{0.734}$ & 0.762 &
$\mathbf{0.770}$ & 0.704 & $\mathbf{0.711}$& 0.770 & 0.759 & 0.883 &
0.840\\
Clayton Kershaw & 0.738 & 0.719 & 0.716 & 0.667 & 0.909 & $\mathbf
{1.125}$ & 0.710 & 0.691 & 0.996 & 0.800 & 0.757 & $\mathbf{0.837}$\\
Chris Carpenter & 0.751 & 0.640 & 0.691 & $\mathbf{0.732}$ & 0.757 &
0.692 & 0.753 & $\mathbf{0.791}$& 0.667 & $\mathbf{0.944}$ & 0.809 &
$\mathbf{0.914}$\\
Matt Garza & 0.827 & 0.792 & 0.794 & $\mathbf{0.794}$ & 0.725 & 0.622 &
0.779 & $\mathbf{0.834}$& 0.692 & 0.622 & 0.797 & $\mathbf{0.880}$\\
Adam Wainwright & 0.689 & $\mathbf{0.690}$ & 0.692 & 0.648 & 0.609 &
0.578 & 0.694 & $\mathbf{0.738}$& 0.554 & $\mathbf{0.636}$ & 0.707 &
0.705\\
Ubaldo Jimenez & 0.735 & 0.709 & 0.767 & $\mathbf{0.771}$ & 0.867 &
0.850 & 0.751 & 0.741& 0.881 & $\mathbf{0.968}$ & 0.741 & $\mathbf
{0.755}$\\
Matt Cain & 0.753 & $\mathbf{0.878}$ & 0.665 & 0.610 & 0.828 & $\mathbf
{0.897}$ & 0.693 & 0.610& 0.815 & $\mathbf{0.854}$ & 0.755 & $\mathbf
{0.774}$\\
Jonathan Sanchez & 0.894 & 0.842 & 0.850 & 0.805 & 0.794 & 0.672 &
0.842 & $\mathbf{1.096}$& 0.791 & 0.672 & 0.968 & $\mathbf{1.165}$\\
Roy Oswalt & 0.710 & 0.639 & 0.787 & 0.750 & 0.876 & 0.667 & 0.777 &
0.651& 1.022 & 0.967 & 0.687 & 0.666\\
Justin Verlander & 0.769 & 0.763 & 0.757 & $\mathbf{0.758}$ & 0.883 &
0.758 & 0.758 & 0.755& 0.930 & 0.914 & 0.758 & $\mathbf{0.776}$\\
Josh Johnson & 0.683 & 0.661 & 0.673 & $\mathbf{0.701}$ & 0.716 & 0.691
& 0.689 & $\mathbf{0.755}$& 0.658 & $\mathbf{0.691}$ & 0.684 & $\mathbf
{0.764}$\\
John Danks & 0.813 & $\mathbf{0.845}$ & 0.807 & $\mathbf{0.820}$ &
0.772 & 0.692 & 0.812 & $\mathbf{0.868}$& 0.758 & $\mathbf{0.855}$ &
0.819 & $\mathbf{0.877}$\\
Edwin Jackson & 0.916 & $\mathbf{1.053}$ & 0.854 & 0.725 & 0.948 &
0.928 & 0.840 & 0.802 & 0.987 & 0.930 & 0.966 & 0.929\\
Max Scherzer & 0.843 & 0.707 & 0.794 & 0.732 & 0.854 & 0.728 & 0.789 &
$\mathbf{0.804}$& 0.847 & $\mathbf{0.899}$ & 0.868 & 0.774\\
Ted Lilly & 0.714 & 0.568 & 0.740 & 0.666 & 0.774 & 0.770 & 0.689 &
0.613 & 0.792 & 0.780 & 0.708 & 0.690\\
\hline
\end{tabular*}}}
\end{sidewaystable}

\begin{sidewaystable}
\tablewidth=\textwidth
\caption{Values obtained from CRLIB
after training on one year of data and testing on another. Superscripts
$\pi^p, \pi^g$ denote the expected
reward from the $\{0,0\}$ state when following the pitcher-specific or
general batting strategy. The bolded
pitcher-specific batting strategies are batting strategies that exploit
the respective pitcher on
the respective training and test dataset pair}\label{fig:cRLIBres}
{\fontsize{8.3}{10.3}{\selectfont
\begin{tabular*}{\textwidth}{@{\extracolsep{\fill}}lcccccccccccc@{}}
\hline
& \multicolumn{4}{c}{\textbf{Training on 2008 data}} &
\multicolumn{4}{c}{\textbf{Training on 2009 data}} &
\multicolumn{4}{c@{}}{\textbf{Training on 2010 data}} \\[-6pt]
& \multicolumn{4}{c}{\hrulefill} &
\multicolumn{4}{c}{\hrulefill} &
\multicolumn{4}{c@{}}{\hrulefill} \\
 & \multicolumn{2}{c}{\textbf{2009}} & \multicolumn{2}{c}{\textbf{2010}} &
\multicolumn{2}{c}{\textbf{2008}} & \multicolumn{2}{c}{\textbf{2010}} & \multicolumn
{2}{c}{\textbf{2008}} & \multicolumn{2}{c}{\textbf{2009}} \\[-6pt]
 & \multicolumn{2}{c}{\hrulefill} & \multicolumn{2}{c}{\hrulefill} &
\multicolumn{2}{c}{\hrulefill} & \multicolumn{2}{c}{\hrulefill} & \multicolumn
{2}{c}{\hrulefill} & \multicolumn{2}{c@{}}{\hrulefill} \\
\textbf{Player}& \multicolumn{1}{c}{$\bolds{J^{\pi^g}}$} &
\multicolumn{1}{c}{$\bolds{J^{\pi^p}}$} &
\multicolumn{1}{c}{$\bolds{J^{\pi^g}}$} &
\multicolumn{1}{c}{$\bolds{J^{\pi^p}}$} &
\multicolumn{1}{c}{$\bolds{J^{\pi^g}}$} &
\multicolumn{1}{c}{$\bolds{J^{\pi^p}}$} &
\multicolumn{1}{c}{$\bolds{J^{\pi^g}}$} &
\multicolumn{1}{c}{$\bolds{J^{\pi^p}}$}
&\multicolumn{1}{c}{$\bolds{J^{\pi^g}}$} &
\multicolumn{1}{c}{$\bolds{J^{\pi^p}}$} &
\multicolumn{1}{c}{$\bolds{J^{\pi^g}}$} &
\multicolumn{1}{c}{$\bolds{J^{\pi^p}}$}\\
\hline
Roy Halladay & 0.644 & 0.594 & 0.603 & 0.585 & 0.544 & $\mathbf{0.575}$
& 0.642 & $\mathbf{0.784}$& 0.538 & $\mathbf{0.558}$ & 0.568 & $\mathbf
{0.765}$\\
Cliff Lee & 0.564 & 0.534 & 0.532 & $\mathbf{0.595}$ & 0.600 & 0.562 &
0.522 & 0.478 & 0.627 & $\mathbf{0.700}$ & 0.536 & $\mathbf{0.557}$\\
Cole Hamels & 0.644 & $\mathbf{0.701}$ & 0.622 & $\mathbf{0.664}$ &
0.540 & 0.537 & 0.595 & 0.484& 0.526 & $\mathbf{0.634}$ & 0.653 &
0.617\\
Jon Lester & 0.625 & 0.587 & 0.651 & $\mathbf{0.720}$ & 0.547 & 0.527 &
0.600 & 0.531& 0.556 & $\mathbf{0.588}$ & 0.570 & $\mathbf{0.610}$\\
Zack Greinke & 0.526 & 0.465 & 0.630 & $\mathbf{0.668}$ & 0.632 &
$\mathbf{0.643}$ & 0.636 & $\mathbf{0.637}$& 0.658 & $\mathbf{0.762}$ &
0.499 & $\mathbf{0.559}$\\
Tim Lincecum & 0.531 & $\mathbf{0.555}$ & 0.657 & 0.497 & 0.590 &
$\mathbf{0.650}$ & 0.565 & $\mathbf{0.580}$ & 0.578 & $\mathbf{0.583}$
& 0.543 & $\mathbf{0.545}$\\
CC Sabathia & 0.602 & 0.543 & 0.604 & $\mathbf{0.604}$ & 0.514 &
$\mathbf{0.657}$ & 0.546 & $\mathbf{0.659}$& 0.458 & $\mathbf{0.581}$ &
0.588 & $\mathbf{0.615}$\\
Johan Santana & 0.569 & $\mathbf{0.591}$ & 0.510 & $\mathbf{0.533}$ &
0.538 & $\mathbf{0.650}$ & 0.526 & $\mathbf{0.533}$& 0.482 & $\mathbf
{0.575}$ & 0.519 & $\mathbf{0.604}$\\
Felix Hernandez & 0.580 & $\mathbf{0.607}$ & 0.523 & $\mathbf{0.564}$ &
0.602 & 0.551 & 0.529 & $\mathbf{0.565}$ & 0.526 & 0.504 & 0.577 &
0.556\\
Chad Billingsley & 0.581 & $\mathbf{0.597}$ & 0.608 & 0.576 & 0.565 &
$\mathbf{0.573}$ & 0.598 & $\mathbf{0.626}$& 0.513 & 0.490 & 0.532 &
0.463\\
Jered Weaver & 0.612 & 0.590 & 0.516 & $\mathbf{0.555}$ & 0.554 &
$\mathbf{0.575}$ & 0.497 & $\mathbf{0.634}$& 0.573 & 0.546 & 0.638 &
$\mathbf{0.666}$\\
Clayton Kershaw & 0.452 & 0.318 & 0.527 & 0.477 & 0.733 & 0.435 & 0.506
& $\mathbf{0.536}$& 0.812 & 0.506 & 0.390 & $\mathbf{0.543}$\\
Chris Carpenter & 0.542 & 0.508 & 0.638 & 0.506 & 0.459 & 0.000 & 0.686
& $\mathbf{0.882}$& 0.288 & 0.269 & 0.597 & $\mathbf{0.640}$\\
Matt Garza & 0.641 & 0.639 & 0.611 & $\mathbf{0.647}$ & 0.539 & 0.500 &
0.663 & $\mathbf{0.670}$ & 0.574 & $\mathbf{0.589}$ & 0.597 & $\mathbf
{0.631}$\\
Adam Wainwright & 0.608 & 0.590 & 0.595 & 0.527 & 0.620 & $\mathbf
{0.785}$ & 0.589 & 0.588& 0.464 & $\mathbf{0.504}$ & 0.601 & 0.546\\
Ubaldo Jimenez & 0.547 & $\mathbf{0.636}$ & 0.529 & $\mathbf{0.565}$ &
0.597 & $\mathbf{0.690}$ & 0.497 & $\mathbf{0.632}$& 0.604 & 0.560 &
0.594 & $\mathbf{0.597}$\\
Matt Cain & 0.586 & $\mathbf{0.658}$ & 0.478 & $\mathbf{0.521}$ & 0.600
& 0.593 & 0.552 & 0.463& 0.491 & $\mathbf{0.554}$ & 0.575 & $\mathbf
{0.597}$\\
Jonathan Sanchez & 0.730 & $\mathbf{0.730}$ & 0.585 & 0.543 & 0.590 &
0.478 & 0.558 & $\mathbf{0.773}$& 0.593 & 0.447 & 0.729 & $\mathbf
{0.861}$\\
Roy Oswalt & 0.621 & 0.534 & 0.498 & $\mathbf{0.510}$ & 0.564 & 0.453 &
0.471 & $\mathbf{0.473}$& 0.598 & $\mathbf{0.646}$ & 0.588 & 0.542\\
Justin Verlander & 0.601 & 0.493 & 0.524 & 0.480 & 0.597 & 0.502 &
0.518 & $\mathbf{0.541}$& 0.619 & 0.597 & 0.547 & 0.546\\
Josh Johnson & 0.519 & $\mathbf{0.631}$ & 0.472 & $\mathbf{0.520}$ &
0.667 & $\mathbf{0.762}$ & 0.505 & $\mathbf{0.519}$& 0.582 & $\mathbf
{0.594}$ & 0.554 & 0.544\\
John Danks & 0.682 & $\mathbf{0.749}$ & 0.537 & 0.492 & 0.518 & 0.512 &
0.502 & 0.478& 0.502 &$\mathbf{0.538}$ & 0.642 & $\mathbf{0.656}$\\
Edwin Jackson & 0.659 & $\mathbf{0.886}$ & 0.657 & 0.635 & 0.712 &
$\mathbf{0.785}$ & 0.617 & 0.575 & 0.708 & $\mathbf{0.758}$ & 0.750 &
0.692\\
Max Scherzer & 0.667 & 0.585 & 0.533 & $\mathbf{0.634}$ & 0.415 & 0.326
& 0.536 & 0.516 & 0.401 & $\mathbf{0.537}$ & 0.690 & $\mathbf{0.710}$\\
Ted Lilly & 0.547 & $\mathbf{0.614}$ & 0.624 & $\mathbf{0.636}$ & 0.592
& 0.516 & 0.589 & 0.522& 0.576 & 0.507 & 0.497 & 0.429\\
\hline
\end{tabular*}}}
\end{sidewaystable}

It is apparent that no relationship exists between the performance
of the pitcher-specific batting strategy and the train/test dataset
pair that it was computed and evaluated on. This shows that the
exploited pitchers' pitcher-specific batting strategies do not rely on
seasonal statistics from either the training or test data. Instead,
these batting strategies rely on the pitcher's decision-making, which
is presumably reflected through the pitcher's pitch selection at each
pitch count. It follows that these characteristics are not reflected in
the respective pitcher's seasonal statistics.

The inclusion of pitch-types in the CRLIB model resulted in a
larger number of exploited pitchers than SRLIB. This suggests the
degree to which a pitcher's pitch selection is influenced by the pitch
count. We use the spatial component to simulate the batter's
performance against an exploited pitcher when using the
pitcher-specific batting strategy, which illustrates utility of these
batting strategies in actual players' decision-making.


\subsection{Simulating batting strategies with the spatial component}

We chose Miguel Cabrera, Joey Votto, Joe Mauer, Ichiro Suzuki,
Jose Bautista, Adrian Gonzalez, Carl Crawford, Matt Holliday, Manny
Ramirez and Alex Rodriguez as our elite batters, and Derek Jeter, Ryan
Howard, Mark Teixeira, Nick Markakis, Brandon Philips, Carlos Gonzalez,
Prince Fielder, Matt Kemp, Evan Longoria and Troy Tulowitzki as our
nonelite batters.

We used CRLIB's strategies from the 2010/2009 train/test dataset pair,
where each batter's $\alpha$ is calculated from the 2010 season, which
is used to construct the respective batter's believed trajectories on
the 2009 data (for the respective pitcher). The actual and simulated
statistics accrued by the elite batters are provided in Tables \ref{fig:actualbatstats9} and \ref{fig:rez09}, respectively.

\begin{table}
\caption{The actual statistics of the elite batters when
facing the respective pitcher in the 2009
season\protect\tabnoteref[*]{ttl6}}\label{fig:actualbatstats9}
\begin{tabular*}{\textwidth}{@{\extracolsep{\fill}}lccccccccc@{}}
\hline
\textbf{Batter} & \textbf{Pitcher} & \textbf{PA} & \textbf{AB} & \textbf{H} &
\textbf{BB} & \textbf{SO} & \textbf{AVG} & \textbf{OBP} & \textbf{SLG} \\
\hline
Miguel Cabrera & Zack Greinke & 14 & 14 & 2 & 0 & 5 & 0.143 & 0.143 &
0.286 \\
Joey Votto & Roy Oswalt & 12 & 12 & 4 & 0 & 2 & 0.333 & 0.333 & 0.500
\\
Joe Mauer & Justin Verlander & 14 & 12 & 4 & 2 & 3 & 0.333 & 0.429 &
0.917 \\
Ichiro Suzuki & CC Sabathia & 12 & 10 & 4 & 2 & 3 & 0.400 & 0.500 &
0.600 \\
Jose Bautista & Jon Lester & 11 & \phantom{0}8 & 1 & 3 & 2 & 0.125 & 0.364 &
0.125 \\
{Adrian Gonzalez} & {Matt Cain} &
\phantom{0}7 & \phantom{0}7 & 6 & 0 & 1 & 0.857 & 0.857 & 1.714 \\
Carl Crawford & CC Sabathia & 13 & 13 & 4 & 0 & 4 & 0.308 & 0.308 &
0.538 \\
Matt Holliday & Clayton Kershaw & \phantom{0}8 & \phantom{0}5 & 2 & 3 & 2 & 0.400 & 0.625 &
1.025 \\
Manny Ramirez & Jonathan Sanchez & 11 & \phantom{0}9 & 5 & 2 & 1 & 0.556 & 0.636
& 1.111 \\
Alex Rodriguez & Roy Halladay & 18 & 17 & 6 & 1 & 2 & 0.353 & 0.389 &
0.471 \\
\hline
\end{tabular*}
\tabnotetext[*]{ttl6}{Please see \hyperref[sect:lingo]{Appendix} for
baseball terminology.}
\end{table}

\begin{table}
\caption{Simulated performance of the elite batters on
the 2009 season using the strategic and spatial components, both of
which are trained on 2010 data\protect\tabnoteref[*]{ttl7}}\label{fig:rez09}
\begin{tabular*}{\textwidth}{@{\extracolsep{\fill}}lcccccccc@{}}
\hline
\textbf{Batter} & \textbf{Pitcher} & \textbf{AB} & \textbf{H} &
\textbf{BB} & \textbf{SO} & \textbf{AVG} & \textbf{OBP} & \textbf{SLG} \\
\hline
Miguel Cabrera & Zack Greinke & 911 & 295 & 0 & \phantom{0}93 & 0.324 & 0.324 &
0.538 \\
Joey Votto & Roy Oswalt & 655 & 136 & 0 & 200 & 0.208 & 0.208 & 0.379
\\
Joe Mauer & Justin Verlander & 723 & 169 & 0 & 127 & 0.234 & 0.234 &
0.390 \\
Ichiro Suzuki & CC Sabathia & 775 & 237 & 0 & 138 & 0.306 & 0.306 &
0.452 \\
Jose Bautista & Jon Lester & 701 & 198 & 0 & 181 & 0.282 & 0.282 &
0.413 \\
Adrian Gonzalez & Matt Cain & 136 & \phantom{0}39 & 0 & \phantom{00}0 & 0.287 & 0.287 & 0.515
\\
Carl Crawford & CC Sabathia & 978 & 280 & 0 & 231 & 0.286 & 0.286 &
0.457 \\
Matt Holliday & Clayton Kershaw & 755 & 187 & 0 & 158 & 0.247 & 0.247
& 0.358 \\
Manny Ramirez & Jonathan Sanchez & 635 & 321 & 0 & \phantom{0}93 & 0.506 & 0.506
& 0.624 \\
Alex Rodriguez & Roy Halladay & 662 & 234 & 0 & \phantom{0}81 & 0.353 & 0.353 &
0.523 \\
\hline
\end{tabular*}
\tabnotetext[*]{ttl7}{Please see \hyperref[sect:lingo]{Appendix} for
baseball terminology.}
\end{table}

\begin{table}[b]
\caption{The actual statistics of the nonelite batters
when facing the respective pitcher in the 2009 season\protect\tabnoteref[*]{ttl8}}\label{fig:stactualbatstats9}
\begin{tabular*}{\textwidth}{@{\extracolsep{\fill}}lccccccccc@{}}
\hline
\textbf{Batter} & \textbf{Pitcher} & \textbf{PA} &
\textbf{AB} & \textbf{H} & \textbf{BB} & \textbf{SO} & \textbf{AVG} &
\textbf{OBP} & \textbf{SLG} \\
\hline
Derek Jeter & Matt Garza & 14 & 14 & 3 & 0 & 2 & 0.214 & 0.214 & 0.357
\\
Ryan Howard & Tim Lincecum & \phantom{0}7 & \phantom{0}7 & 2 & 0 & 4 & 0.286 & 0.286 & 0.429
\\
Mark Teixeira & Cliff Lee & 14 & 13 & 2 & 0 & 3 & 0.154 & 0.214 &
0.231 \\
Nick Markakis & Jon Lester & 13 & 13 & 1 & 0 & 6 & 0.077 & 0.077 &
0.077 \\
Brandon Philips & Chris Carpenter & 11 & 10 & 1 & 3 & 2 & 0.100 &
0.182 & 0.282 \\
Carlos Gonzalez & Matt Cain & 14 & 13 & 3 & 1 & 4 & 0.231 & 0.286 &
0.231 \\
Troy Tulowitzki & Clayton Kershaw & 14 & 13 & 0 & 1 & 6 & 0.000 &
0.071 & 0.000 \\
Matt Kemp & Jonathan Sanchez & 11 & 11 & 3 & 0 & 0 & 0.273 & 0.273 &
0.364 \\
Evan Longoria & Roy Halladay & 23 & 17 & 4 & 4 & 4 & 0.235 & 0.348 &
0.235 \\
Prince Fielder & Matt Cain & \phantom{0}9 & \phantom{0}7 & 1 & 1 & 1 & 0.143 & 0.222 & 0.286
\\
\hline
\end{tabular*}
\tabnotetext[*]{ttl8}{Please see \hyperref[sect:lingo]{Appendix} for
baseball terminology.}
\end{table}

\begin{table}
\caption{Simulated performance of the nonelite batters on
the 2009 season using the spatial and strategic components, both of
which are trained on 2010 data\protect\tabnoteref[*]{ttl9}}\label{fig:strez09}
\begin{tabular*}{\textwidth}{@{\extracolsep{\fill}}lcccccccc@{}}
\hline
\textbf{Batter} & \textbf{Pitcher} & \textbf{AB} &
 \textbf{H} & \textbf{BB} & \textbf{SO} & \textbf{AVG} & \textbf{OBP} & \textbf{SLG} \\
\hline
Derek Jeter & Matt Garza & \phantom{0}257 & 110 & \phantom{00}0 & \phantom{00}0 & 0.428 & 0.428 & 0.656 \\
Ryan Howard & Tim Lincecum & \phantom{0}487 & 137 & \phantom{00}0 & 131 & 0.281 & 0.281 &
0.413 \\
Mark Teixeira & Cliff Lee & \phantom{0}301 & \phantom{0}93 & \phantom{00}0 & \phantom{0}66 & 0.309 & 0.309 & 0.495
\\
Nick Markakis & Jon Lester & \phantom{0}925 & 383 & 100 & 125 & 0.415 & 0.522 &
0.520 \\
Brandon Philips & Chris Carpenter & \phantom{0}269 & 161 & \phantom{00}0 & \phantom{00}0 & 0.599 & 0.599
& 1.115 \\
Carlos Gonzalez & Matt Cain & \phantom{0}283 & \phantom{0}86 & \phantom{00}0 & \phantom{0}59 & 0.304 & 0.304 &
0.435 \\
Troy Tulowitzki & Clayton Kershaw & 1017 & 216 & \phantom{00}0 & 176 & 0.212 &
0.212 & 0.306 \\
Matt Kemp & Jonathan Sanchez & \phantom{0}602 & 166 & \phantom{00}0 & 261 & 0.276 & 0.276 &
0.372 \\
Evan Longoria & Roy Halladay & 1309 & 429 & \phantom{00}0 & 153 & 0.328 & 0.328 &
0.503 \\
Prince Fielder & Matt Cain & \phantom{0}450 & 124 & \phantom{00}0 & 115 & 0.276 & 0.276 &
0.429 \\
\hline
\end{tabular*}
\tabnotetext[*]{ttl9}{Please see \hyperref[sect:lingo]{Appendix} for
baseball terminology.}
\end{table}

When comparing the elite batters' simulated statistics to their
actual statistics on the 2009 test data's at-bats, the simulated
statistics do not show an appreciable performance improvement in
comparison to the actual statistics. We evaluated the performance of
the ten nonelite batters to preclude the possibility that an exploited
pitcher's pitcher-specific batting strategy is detrimental to an elite
hitter's ability to get on base. These nonelite batters are of course
exceptional hitters, but are not considered as elite at other aspects
of the game. If following an exploited pitcher's pitcher-specific
batting strategy is detrimental to elite hitters, then nonelite hitters
may still benefit.

Comparing the nonelite batters' actual statistics to their
simulated statistics, shown in Tables \ref{fig:stactualbatstats9}
and \ref{fig:strez09}, respectively, the simulated statistics are
superior to the actual statistics for all of the batters. This suggests
that the exploited pitchers' pitcher-specific batting strategies are
beneficial for nonelite players. Considering that Joe Mauer, Ichiro
Suzuki, Joey Votto, Manny Ramirez and Alex Rodriguez are generational
talents, it is possible that elite batters make atypical decisions that
only they can benefit from.\looseness=1

\subsection{Discussion}
Before discussing the limiting factors on the experiment, we briefly
mention that the batter's decision-making can be exploited in a similar
manner: Batter-specific behaviour could be modeled as a stochastic
process. Reinforcement Learning could then be used to obtain optimal
pitching strategies against both specific batters and the general
population of batters. This idea is relegated to future
work.

\subsubsection{Potential limitations: Strategic component}\label{sect:stratdisc}

The reward function gives a higher reward for a single than a
walk, reflecting the idea that a batter should not be indifferent
between a walk and a single. This is because a single advances
baserunners that are not on first base, whereas a walk does not. We
acknowledge that there are cases, such as when there are no baserunners
or one baserunner on first base, where a single should be considered
equivalent to a walk. However, it is desirable to compute a batting
strategy that maximizes a batter's expectation of reaching base while
also trying to win the game; advancing runners is critical to winning
in baseball. In contrast, the slugging percentage (SLG) calculation can
be interpreted as a reward function that quantifies the batter's
preference of a single, double, triple and home run as 1, 2, 3 and 4,
while ignoring walks. Given that the On-base Plus Slugging (OPS) metric
is often used to measure a player's ability, but does not consider
walks and hits as a function of the batting action, it was felt that
the reward function used in the study must address this shortcoming by
giving a reward of 1 for a walk; by doing so, all outcomes are
considered by the same reward function, where the lowest nonzero reward
is a walk, which addresses the shortcomings of OPS.

It is possible that performance can be improved by tuning the
reward function through Inverse Reinforcement Learning (IRL) [\citet
{irl}], which can recover the optimal reward function. One caveat with
IRL, however, is that it requires a near-optimal policy to recover the
optimal reward function, which suggests that a good reward function is
first required to develop a near-optimal policy. We therefore see an
interdependence between the optimal policy and reward function, where
addressing this interdependence is a focal point of Reinforcement
Learning research.

Under the assumption that all pitchers are \textit{not} equally
exploitable, the results show that strategizing against a specific
pitcher is statistically better than strategizing against a group of
the pitchers more than 50\% of the time; assuming that pitchers are
\textit{equally} susceptible to being exploited by either strategy, the
hypothesis was rejected. However, we argue the statistically
significant result under the assumption of unequal susceptibility is
stronger because each pitcher has unique behaviour that is reflected in
the transition probabilities in each state of the at-bat (see
Section \ref{sect:comps} for further information). Given that only two
fewer pitchers are exploited under the assumption that pitchers are
equally susceptible to being exploited, which resulted in rejection of
our hypothesis, it is possible that evaluating the hypothesis over a
larger number of pitchers would result in a nonrejection under \textit
{both} assumptions with the CRLIB model.

Important parts of the data provided by MLB's GameDay system were
incomplete. There were cases where the at-bat's pitch sequence
consisted of one (or more) pitches that did not have a
pitch-type. 
For the 610,130 at-bats in the database, there were 32,632
pitches concluded for the at-bat without providing the pitch-type. This
was detrimental to CRLIB because the last pitch of an at-bat determined
the terminal state, and the pitch-type determines the state being
transitioned from. We therefore skipped over pitches with missing
pitch-types and simply incremented the pitch count by observing whether
the unlabeled pitch was a ball or strike. We also observed that the
2008 season of data contained many more pitch-types than the 2009 and
2010 seasons for the same pitcher, which suggested that the GameDay
system data was maturing in its initial years; it's feasible that
future work using newer data will report even better results than those
in this article.

Using the four generalized pitch-types mentioned in Section \ref{desc2} may have a large impact on our results. However, there are 15
different pitch-types in the GameDay system, and it is unreasonable to
give CRLIB $15 \times12 + 6$ states due to issues that would arise
with data sparsity, as we did not want to use any sampling techniques.

A potential criticism of our batting strategy evaluations is that
they are evaluated over an entire season of data. We emphasize the fact
that we are not assuming that a pitcher does not adjust to the
pitcher-specific batting strategy. After initially using the
pitcher-specific batting strategy in the real world, we update the
strategy using online learning algorithms, such as
State--Action--Reward--State--Action (SARSA) [\citet{sb}]. Online learning
algorithms update and recompute the pitcher-specific batting strategy
using the pitch-by-pitch data after the pitcher adjusts to being
exploited by the initial batting strategy. We use a season of
pitch-by-pitch data to show the potential of our approach when the
pitcher does not know they are being exploited. Training and testing
our model over two different seasons of pitch-by-pitch data for the
same pitcher shows that the pitcher-specific batting strategy's
performance is not a consequence of luck.

\subsubsection{Potential limitations: Simulating batting strategies
with spatial component}
Readers may argue that assuming the batter can hit the pitch if it
is identified correctly is not representative of the at-bat setting.
However, we are simulating the at-bat setting in a probabilistic
manner, which means that there is no guarantee that the batting action
will yield a fortuitous outcome. This is because batters reach base
less than being out over an entire season, and this is reflected by the
large probability of the out state.

We do not consider predicting the end location of pitches because
there are two assumptions that we do not agree with: we are assuming
that the pitch-type and current state are related to a pitch's end
location, and we are assuming that the pitch's end location is related
to the pitch-type, neither of which are true. In the real world,
batters rarely know where the pitch is going. Instead, they rely on
identifying the pitch-type prior to making the choice of swinging or
not. The limitations of the spatial component stem from the absence of
the raw spatial information for each pitch.

Another issue some may have with the spatial component is that, in
conjunction with an exploited pitcher's pitcher-specific batting
strategy, our model fails to produce walks in all but one case. The
lack of walks is a result of CRLIB's pitcher-specific strategy
suggesting that the batters swing in pitch counts where there are three
balls. Swinging at pitch counts with three balls maximizes the batter's
expectation because pitchers do not want to walk the batter. We
mentioned that our model puts a higher priority on advancing
baserunners, and it is possible that this prioritization decreased the
OBP and AVG statistics. This decrease is explained by the fact that the
batters are swinging in states with large expected rewards, which can
only be reached with a base hit.

Additionally, the actual number of pitches thrown in the at-bat
may be a limitation on achieving walks in our simulation. For example,
assume that the actual 2009 data shows that the batter chose to swing
in a certain pitch count, and this swing led to them being out. Let us
also assume that exploited pitcher's pitcher-specific batting strategy
selects the batting action ``stand.'' If this pitch is not thrown for a
strike, the at-bat is incomplete because there are no more pitches
left. A similar limitation that arises is when the batter identifies a
pitch at the respective pitch count in the test data, but the training
data did not contain a case with this pitch count and pitch-type. For
example, the batter identifies the pitch as a fastball in a $\{3,0\}$
count, but the training data only contained curveballs being thrown
from the $\{3,0\}$ count. In either of these situations, we skip the
at-bat.\footnote{We define ``skipped'' as the termination of the current
simulation. This at-bat is only simulated again if the 100 simulation
limit has not been exceeded.} These examples illustrate how real-world
data can be limiting on a simulation.

\subsubsection{Reinforcement Learning in football}\label{sect:ndpdisc}
Patek and Bertsekas (\citeyear{ndp}) used Reinforcement Learning to simulate the offensive play
calling for a simplified version of American Football. They computed an
optimal policy for the offensive team by giving generated sample data
(that was representative of typical play) as input to their model.

The optimal policy suggested running, passing and running plays
when the distance from line of scrimmage and ``our'' own goal line was
between 1 and 65 yards, 66 yards and 94 yards, and 95 and 100
(touchdown) yards. The optimal policy produced an expected reward of
$-0.9449$ points when starting from the twenty yard line. This meant that
if the ``our'' team received the ball at the twenty yard line every
time, they would lose the game. It is possible that the result could
have produced a positive expected reward if real-world data was used,
but this data was not available at the time.

The appealing property of Reinforcement Learning is that it allows
the evaluation of arbitrary strategies given as input to the Policy
Evaluation algorithm. One strategy that was reflective of good
play-calling in football produced a slightly worse expected reward
($-1.27$ points) than the optimal policy. Considering that this strategy
was intuitive, manually constructed and did not perform much worse than
the optimal strategy, Reinforcement Learning provides a platform for
investigating the strategic aspects in sports. After all, every team's
goal is to devise a strategy that maximizes the team's opportunity to
win the game.


Reinforcement Learning's applicability to baseball is tantalizing
because team performance depends on individual performance. The
distinguishing aspect of Reinforcement Learning in baseball (from
football) is the scope of the strategy: in football, a policy
reflective of ``good'' play-calling does not change significantly with
the opponent. In baseball, the batting strategy changes significantly
with the opponent, as the batting strategy is pitcher-specific. Using
an exploited pitcher's pitcher-specific batting strategy against the
respective pitcher should increase the team's opportunity of winning
the game, as it increases the batter's expectation of reaching base.

\section{Conclusion}\label{conc}
This article shows how Reinforcement Learning algorithms [\citet
{ndp,ndpref,sb}] can be applied to Markov Decision Processes [\citet
{lawler}] to statistically analyze baseball's at-bat setting. With the
wealth of information provided at the pitch-by-pitch level, evaluating
a player's decision-making ability is no longer unrealistic.

Earlier work alluded to the amount of information contained in
real-world baseball data being a limiting factor for analysis
[\citet{mcb} and \citet{oera}]. Noting this,
what is particularly impressive about CRLIB's statistically
significant result, which is produced under the assumption that
pitchers are \textit{not} equally susceptible to being exploited (See
Section \ref{sect:stratdisc}), is that each pitcher-specific batting
strategy was computed using a few thousand samples.\vadjust{\goodbreak} In many previous
baseball articles, authors have used Markov Chain Monte Carlo (MCMC)
techniques to produce a sufficient number of samples to test their
model [\citet{breakdown,bayes,bask}]. Currently, this is the only model that
uses only the real-world pitch-by-pitch data to produce a result that
is both intuitive and supported with algorithmic statistics.

Baseball has a rich history of elite batters making decisions that
nonelite batters cannot. This is exemplified by Vladimir Guerrero, a
future first-ballot hall of fame ``bad-ball hitter,'' swinging at a
pitch that bounced off the dirt for a base hit during an MLB game. It
is therefore possible that elite players do not benefit from an
exploited pitcher's pitcher-specific batting strategy because they make
instinctual decisions. Given that the nonelite batters' simulated
statistics are superior to their actual statistics, using the exploited
pitchers' pitcher-specific batting strategies would be incredibly
useful in a real MLB game. We feel that modeling the at-bat setting as
a Markov Decision Process to statistically analyze baseball players'
decision-making will change talent evaluation at the professional level.

%
%
%
%

\begin{appendix}\label{app}
\section*{Appendix: Baseball definitions}\label{sect:lingo}
%
\begin{enumerate}
\item[] A \textit{Plate Appearance \textup{(}PA\textup{)}} is when the batter has
completed their turn batting.
\item[] An \textit{at-bat \textup{(}AB\textup{)}} is when a player takes their turn to
bat against the opposition's pitcher. Unlike Plate Appearances, at-bats
do not include sacrifice flies, walks, being hit by a pitch or
interference by a catcher.
\item[] A \textit{Hit \textup{(}H\textup{)}} is when a player reaches base by putting
the ball into play.
\item[] A \textit{Walk \textup{(}BB\textup{)}} is when a player reaches base on four
balls thrown by the pitcher.
\item[] The [\textit{pitch}] \textit{count} is the current ``state'' of the
at-bat. This is quantified by the number of balls and strikes, denoted
by $B$ and $S$, respectively, thrown by the pitcher.

This is numerically represented as B--S, where $0 \leq B \leq4$, $0
\leq S \leq3$ and $B,S \in\mathbb{Z}_+$, where $\mathbb{Z}_+$ is the
set of $\mathit{positive}$ integers. Note that an at-bat has ended if
any one of the conditions is true:
\begin{itemize}
\item$(B=4)\cap(S < 3)$ (referred to as a walk)
\item$(B < 4) \cap(S=3)$ (a strikeout)
\item$(B<4) \cap(S<3)$ \textit{and} the batter hit the ball inside
the field of play. This results in two disjoint outcomes for the
batter: out or hit. There is a special other case we mention below.
\end{itemize}
\item[] A \textit{Foul} is when a player hits the ball outside the
field of play. If $S < 2$, then the foul is counted as a strike. When
$S=2$ and the next pitch results in a foul, then $S = 2$. That is, if a
ball is hit for a foul with two strikes, the count remains at two
strikes. However, if a foul ball is caught by an opposing player before
it hits the ground, the batter is out regardless of the count.
\item[] A \textit{Batter's count} is defined as a pitch count that
favors the batter---that is, the number of balls are greater than the
number of strikes (exception of the 3--2 count, which is referred to as
a full count).
%
\item[] A \textit{Baserunner} is a player who is on base during his
team's at-bat.
\item[] A \textit{Runner in Scoring Position (RISP)} is a baserunner
who is on second or third base. This is because when an at-bat results
in more than a double, the baserunner on second base can score.
\item[] \textit{Walks plus hits per inning pitched (WHIP)} is a measurement of
how many baserunners a pitcher allows per inning. It is the sum of the
total number of walks and hits divided by the number of innings pitched.
\item[] \textit{Earned Run Average (ERA)} is the average number of earned runs a
pitcher has given up for every nine innings pitched.
\item[] \textit{On Base Percentage (OBP)} is a statistic which represents the
number of times a player reaches base either through a walk or hit when
divided by their total number of at-bats.
\item[] \textit{Slugging Percentage (SLG)} is a measure of the hitter's power.
It assigns rewards 1, 2, 3 and 4 for a single, double, triple and home
run. SLG is a weighted average of these outcomes.
%
\item[] \textit{On-base Plus Slugging (OPS)} is the sum of the On-base
Percentage and Slugging Percentage---that is, OPS${} = {}$OBP${} + {}$SLG.
\end{enumerate}
\end{appendix}

\section*{Acknowledgments}\label{acknowl}
The following acknowledgments are strictly on the behalf of
Gagan Sidhu: I would like to dedicate this work to a Statistical
pioneer, Dr. Leo Breiman (1928--2005), whose fiery and objective writing
style allowed the Statistical learning community to garner recognition.
This work would not have been considered statistics without Dr.
Breiman's \textit{Statistical Science} article that delineated
algorithmic and classical statistical methods. Big thanks to Dr.
Michael Bowling (University of Alberta) for both supervising and
mentoring this project in its infancy, Dr. Brian Caffo for selflessly
senior-authoring this paper, and Dr. Paddock, the Associate editor and
reviewers for providing helpful comments. Last, thanks to Lauren Styles
for assisting in marginalizing my posterior distribution.

\begin{supplement}[id=suppA]
\stitle{Supplement to ``MONEYBaRL: Exploiting pitcher decision-making using
Reinforcement Learning''}
\slink[doi]{10.1214/13-AOAS712SUPP} 
\sdatatype{.pdf}
\sfilename{aoas712\_supp.pdf}
\sdescription{A high-level overview of the technical details of the
implementation used in this article.}
\end{supplement}
%


\printaddresses

\end{document}